\documentclass{article} 
\usepackage[preprint]{colm2025_conference}

\usepackage{microtype}
\usepackage{hyperref}
\usepackage{url}
\usepackage{booktabs}
\usepackage{amsmath}
\usepackage{algorithm}
\usepackage{algpseudocode}
\usepackage{graphicx}
\usepackage{lineno}
\usepackage{colortbl,booktabs}
\usepackage{multicol}
\usepackage{multirow}
\usepackage{mdframed}
\usepackage{listings}
\usepackage{xspace}
\usepackage[normalem]{ulem}
\usepackage{wrapfig}
\usepackage{makecell}
\usepackage{xcolor}
\usepackage{arydshln} 
\usepackage{subfigure}
\usepackage{pifont}

\definecolor{darkblue}{rgb}{0, 0, 0.5}
\definecolor{royalblue}{rgb}{0, 0, 0.7}
\definecolor{darkred}{RGB}{166, 32, 23}
\hypersetup{colorlinks=true, citecolor=darkblue, linkcolor=darkblue, urlcolor=darkblue}

\newcommand{\beginalgcolor}[1]{\color{#1}}

\title{Efficient Process Reward Model Training via Active Learning}

\author{
    \textbf{Keyu Duan}$^{1,2}$\thanks{Work done during the internship in Sea AI Lab. Email: \href{mailto:k.duan@u.nus.edu}{k.duan@u.nus.edu}}\quad
    \textbf{Zichen Liu}$^{1,2}$\quad 
    \textbf{Xin Mao}$^2$\quad
    \textbf{Tianyu Pang}$^2$\quad 
    \textbf{Changyu Chen}$^{2,3}$\quad
    \\ \\
    \textbf{Qiguang Chen}\quad
    \textbf{Michael Qizhe Shieh}$^{1}$\thanks{Corresponding authors.}\quad 
    \textbf{Longxu Dou}$^{2\dag}$
    \\ \\
    $^{1}$National University of Singapore \quad
    $^{2}$Sea AI Lab \quad 
    $^{3}$Singapore Management University 
}

\newcommand{\ourmethod}{\textsc{ActPRM}\xspace}
\newcommand{\oursotamodel}{\textsc{ActPRM-X}\xspace}

\begin{document}

\ifcolmsubmission
	\linenumbers
\fi

\maketitle

\begin{abstract}


Process Reward Models (PRMs) provide step-level supervision to large language models (LLMs), but scaling up training data annotation remains challenging for both humans and LLMs. 
To address this limitation, we propose an active learning approach, \ourmethod, which proactively selects the most uncertain samples for training, substantially reducing labeling costs.
During training, we use the PRM to estimate uncertainty after the forward pass, retaining only highly uncertain data. A capable yet costly reasoning model then labels this data. Then we compute the loss w.r.t. the labels and update the PRM’s weights.
We compare \ourmethod vs. vanilla fine-tuning, on a pool-based active learning setting, demonstrating that \ourmethod reduce 50\% annotation, but achieving the comparable or even better performance. 
Beyond annotation efficiency, we further advance the actively trained PRM by filtering over 1M+ math reasoning trajectories with \ourmethod, retaining 60\% of the data.
A subsequent training on this selected dataset yields a new state-of-the-art (SOTA) PRM on ProcessBench (75.0\%) and PRMBench (65.5\%) compared with same sized models
\footnote{The code is available at \href{https://github.com/sail-sg/ActivePRM}{https://github.com/sail-sg/ActivePRM}}.

\end{abstract}

\vspace{-0.3cm}
\section{Introduction}
\vspace{-0.2cm}

\begin{wrapfigure}{rt}{0.5\linewidth}
    \vspace{-15pt}
    \centering
    \includegraphics[width=\linewidth]{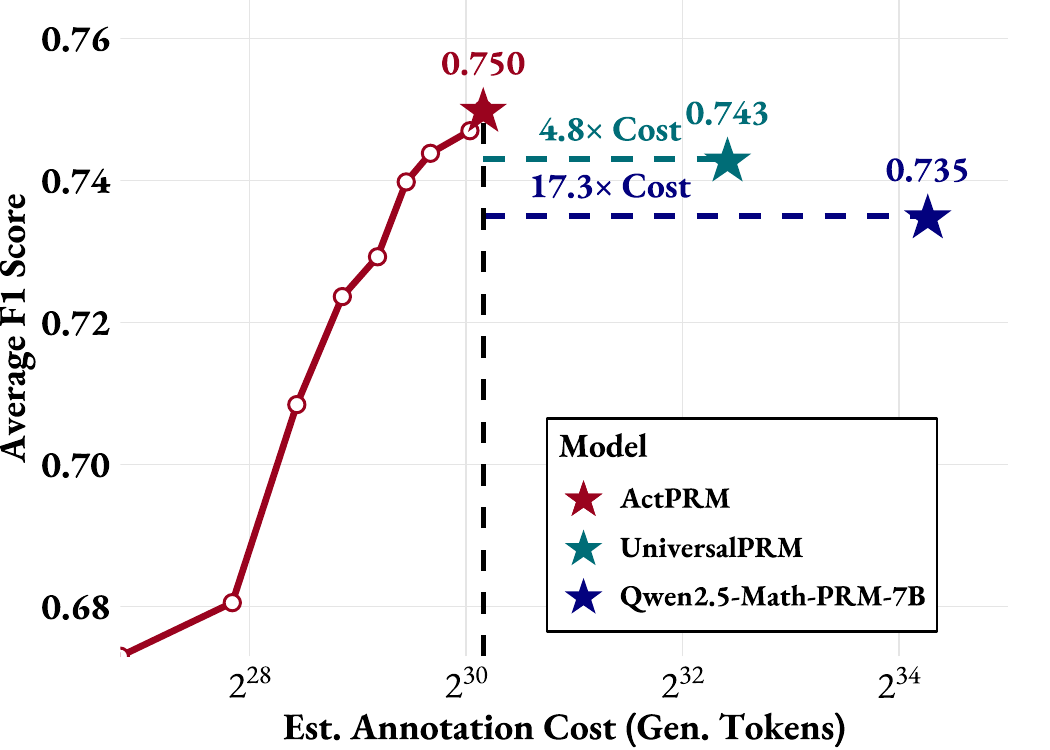}
    \vspace{-0.2cm}
    
    \caption{
    Average F1 score on ProcessBench~\citep{zheng2024processbench} versus estimated annotation cost. 
    \ourmethod outperforms prior SOTA models while requiring merely 20\% of the annotation costs. 
    }
    \label{fig:sec1_performance_vs_cost}
    \vspace{-10pt}
\end{wrapfigure}

Recently, Large Language Models (LLMs)~\citep{deepseekr1, qwen2.5, openai2024openaio1card} have shown remarkable advances in mathematical reasoning, yet they can make mistakes during chain-of-thought (CoT) reasoning despite correct final answers\citep{zheng2024processbench}.
To address this challenge, process reward models~\citep{lightman2023lets,wang2024mathshepherd,zhang2025lessons} were proposed, aiming to identify process errors and provide finer-grained supervision of the training process.

The main challenge in training Process Reward Models (PRMs) lies in obtaining fine-grained step-level annotations, which remain prohibitively expensive. \citet{lightman2023lets} pioneered PRM training by employing human experts to label 75K questions at the step level. While their approach achieved high-quality results (reaching 57.5\% on ProcessBench~\citep{zheng2024processbench}), it does not scale automatically due to the heavy reliance on manual annotation. To reduce human efforts, Monte Carlo (M.C.) Estimation Methods~\citep{wang2024mathshepherd,xiong2024rlhflowprm,luo2024improve} were proposed. However, these approaches come with high computational costs (massive rollouts are required for accurate estimation) and struggle to accurately identify the first error step~\citep{zheng2024processbench}. To address this challenge, Qwen2.5-Math-PRM~\citep{zhang2025lessons} proposed using LLM-as-Judge --- leveraging LLMs to detect the first error step --- and filter out unreliable M.C. labels. 
It significantly boosts the performance of PRM on both ProcessBench~\citep{zheng2024processbench} and PRMBench~\citep{song2025prmbench}.
More recently, UniversalPRM~\citep{tan2025universalprm} relies solely on LLM-as-Judge with ensemble prompting (via majority voting), achieving new SOTA performance on ProcessBench within the same model size. However, the annotation costs are still considerable. We estimate the labeling costs of Qwen2.5-Math-PRM~\citep{zhang2025lessons} and UniversalPRM~\citep{tan2025universalprm} and illustrate them in Figure~\ref{fig:sec1_performance_vs_cost}. It shows that Qwen-Math-PRM-7B and UniversalPRM consume over \(2^{34}\) and \(2^{32}\) generated tokens, respectively.
Refer to Appendix~\ref{sec:labeling_cost_estimation} for estimation strategy.


To reduce annotation costs, we propose \ourmethod, which uses a trained ensemble PRM to identify and select uncertain data for annotation by a high-capability reasoning model.
Our approach trains a PRM with ensemble heads for uncertainty estimation. For each reasoning step, we compute the mean \(\mu\) and standard deviation \(\sigma\) of ensemble predictions, identifying uncertain steps when prediction confidence is outside threshold \(1-\delta_{pred}<\mu<\delta_{pred}\) or variation exceeds \(\delta_{std}\).
We consider a CoT trajectory uncertain \textbf{if any step up to and including the first predicted error meets these criteria}. By annotating only uncertain data and training exclusively on this subset, we significantly reduce labeling costs while maintaining PRM performance.

To validate the effectiveness and efficiency of \ourmethod, we conducted comprehensive experiments in multiple settings:
\begin{itemize}
    \item \textbf{Pool-based Evaluation (Section~\ref{sec:exp_pool_based}):} Using 100K labeled samples, \ourmethod achieved performance comparable to full-data tuning while reducing annotation costs by 50\%. It consistently outperformed random selection under identical budget constraints.
    
    \item \textbf{One-shot Active Learning (Section~\ref{sec:exp_sota}):} Starting with our pool-based model, we applied \ourmethod to select uncertain samples from 1M+ unlabeled CoT trajectories from NuminaMath~\citep{numinamath}. After annotation and fine-tuning, we achieved new SOTA performance of 75.0\% on ProcessBench. As shown in Figure~\ref{fig:sec1_performance_vs_cost}, \ourmethod surpasses prior SOTA models with significantly lower costs—outperforming UniversalPRM~\citep{tan2025universalprm} by 0.7\% using only 20\% of its annotation budget and exceeding Qwen2.5-Math-PRM-7B by 1.5\% with just 6\% of its annotation budget.
\end{itemize}

Our contributions are summarized as follows: 
\ding{182} We propose an uncertainty-aware active learning approach \ourmethod for PRM training that selectively annotates informative reasoning steps using ensemble-based uncertainty estimation, significantly reducing labeling costs while maintaining performance.
\ding{183} \ourmethod achieves state-of-the-art results (75.0\% on ProcessBench, 65.5\% on PRMBench) while requiring only 20\% of the annotation budget compared to prior SOTA method UniversalPRM.
\ding{184} We release all trained models, datasets, and code to ensure reproducibility and facilitate community adoption.

\vspace{-0.2cm}

\section{Preliminaries}
\vspace{-0.2cm}

\subsection{Process Reward Models}

\noindent \textbf{Problem Formulation.} Given a math problem \(q\) and a corresponding solution trajectory \(s =
[s_1, s_2, \dots, s_n]\), where \(s_i\) denotes \(i\)-th step., we require a PRM to identify the correctness of each step until a wrong step is identified.
We only label the steps up to and including the first error step following prior works~\citep{lightman2023lets,zheng2024processbench}, since the afterward steps are genuinely difficult to define their correctness. As a
result, in practice the labels for a solution trajectory are either \([1, 1, \dots, 1]\) or \([1, 1, \dots, 0]\).
Then a PRM could be trained using the typical BCE loss:

\begin{equation}\label{equ:prm_loss}
	\mathcal{L}_{BCE}(s, y|\theta) = -\frac{1}{|s|}\sum_1^{|s|}y_i\log(P_\theta(s_i|s_{[:i]}, q)) +
	(1-y_i)\log(1-P_\theta(s_i|s_{[:i]},q)),
\end{equation}
where \(P_\theta\) is the PRM parameterized by \(\theta\) and \(s_{[:i]}\) denotes the steps before \(s_i\).
When using PRM for inference, we set a threshold \(\delta\)
(usually \(0.5\)) to identify the first step that has a correctness probability \(P_\theta(s_i|s_{[:i]}, q)\) less than \(\delta\).

\noindent \textbf{PRM Implementation Details.} A typical PRM is built upon a pretrained generative LLM by replacing the
causal language model head with a binary classification head that outputs the probability of correctness at
corresponding token position. In practice, we solely need the prediction at the end of each reasoning step and thus a
prediction mask is used to mask out the prediction at the other positions.

\subsection{Uncertainty Estimation for Classification}

\noindent
\textbf{Aleatoric Uncertainty.}
As aforementioned, a typical PRM \(P_\theta\) is trained as a binary classification task. The simplest way to measure
the uncertainty for it is to use aleatoric uncertainty~\citep{valdenegrotoro2022deeperlookaleatoricepistemic}:\footnote{For simplicity, we use \(P_\theta(s_i)\) as the
	aleatoric probability for the \(i\)-th step in the solution trajectory, where the full representation is
	\(p_\theta(s_i|s_{[:i]},q)\) as in Equation~\ref{equ:prm_loss}}.
\begin{equation}\label{equ:aleatoric_uncertainty}
	\mbox{Aleatoric Uncertainty} \propto P_\theta(s_i)\log P_\theta(s_i).
\end{equation}

\noindent
\textbf{Epistemic Uncertainty.}
In addition, ensemble of models is also a common way to estimate epistemic uncertainty~\citep{valdenegrotoro2022deeperlookaleatoricepistemic} by quantifying the
disagreement among ensemble models. For example, \citet{liang2022reward, gleave2022uncertainty} use an ensemble of
reward models to estimate uncertainty in preference learning. for process reward modeling, we could leverage the
standard deviation of the ensemble predictions as the uncertainty estimation:
\begin{equation}\label{equ:epistemic_uncertainty}
	\mbox{Epistemic Uncertainty} \propto \mbox{Var}(\{P_\theta(s_i)\}),
\end{equation}
where \(\{P_\theta\}\) is a set of models. It is worth noting that employing an ensemble of heads built upon a shared
backbone is a common strategy to mitigate computational costs. We empirically study the combination of aleatoric and epistemic uncertainty and find that they are complementary to each other. Experimental results are shown in Section~\ref{sec:exp_pool_based}.
\vspace{-0.2cm}

\section{Related Work}\label{sec:related_work}
\vspace{-0.2cm}

\noindent
\textbf{Active Learning and Uncertainty Estimation.}
Active learning has been widely explored in the alignment of LLMs. Several studies adopt an online bandit formulation, leveraging uncertainty-aware reward models (RMs) for active exploration in response selection~\citep{mehta2023sample,dwaracherla2024efficient,liu2024sampleefficientalignmentllms, melo2024deepbayesianactivelearning,gleave2022uncertainty}. For instance, \citet{mehta2023sample} and \citet{dwaracherla2024efficient} use ensemble-based LLM heads to estimate epistemic uncertainty, prioritizing data most informative for preference learning. Similarly, \citet{melo2024deepbayesianactivelearning} propose an acquisition function combining both entropy (aleatoric uncertainty) and epistemic uncertainty. Our work builds on these approaches, empirically evaluating the role of both uncertainty types in the context of process reward modeling. Beyond active learning, ensemble methods—such as those in \citet{coste2024reward}—have also proven effective in mitigating reward hacking \citep{amodei2016concreteproblemsaisafety}.

\noindent
\textbf{Process Reward Models.}
Different from outcome rewards (e.g., verifiable rewards~\citep{deepseekr1}
for mathematical reasoning problems) that assign rewards based on the final outcome, process rewards are assigned based on the intermediate steps of the problem-solving process. For a question and a corresponding solution with several
steps, a PRM provides finer-grained rewards for each step. Til current stage, process reward modeling contains two
categories: \((i)\) \emph{Process Reward as Q-values} and \((ii)\) \emph{Process Reward as Judger}. The former
one~\citep{wang2024mathshepherd,luo2024improve,xiong2024rlhflowprm,li2024qvalueranking} regards the process reward as the Q-values of the steps
that estimate the probability of the policy model to reach the final correct answer. Specifically, they leverage the
policy model that generates the solution steps to do Monte Carlo Estimation for each step. The estimated Q-values are
used as the process rewards. However, recent works~\citep{zhang2025lessons,zheng2024processbench} show that this kind of process reward modeling suffers from identifying the process errors because it highly depends on the policy model and has large bias with the ground truth distribution. In contrast, the latter one~\citep{lightman2023lets,zhang2025lessons} regards the process reward model as a proxy for identifying the intermediate process errors and the corresponding trained model achieves better performance on several benchmarks~\citep{zheng2024processbench, song2025prmbench}. In this work,
we follow the latter one and regard the process reward as a judge that tries to identify the first error steps in the
solutions if any. In addition, there are other works related to PRM. For example, \citet{yuan2024implicitprm} tries to train a PRM with a fashion of outcome reward modeling (ORM). \citet{cheng2025pure,cui2025eurus} proposed RL training frameworks that integrate PRM as finer-grained supervison.
\vspace{-0.2cm}
\section{Efficient Process Reward Labeling via Active Learning} 
\vspace{-0.2cm}

Labeling the process rewards for a large-scale dataset is very expensive as it either requires human experts to annotate the correctness of each step for each solution as in the previous work~\citep{lightman2023lets} or leverages highly capable generative models to imitate human experts~\citep{zhang2025lessons}. Even though the latter one is automated, it is still resource-consuming since the test time scales up with the difficulty of math problems. 

\begin{figure}[t]
	\vspace{-1cm}
	\scalebox{0.95}{
		\begin{minipage}{\linewidth}
			\begin{algorithm}[H]
				\renewcommand{\algorithmicrequire}{\textbf{Input:}}
				\renewcommand{\algorithmicensure}{\textbf{Output:}}
				\caption{PRM Active Learning with Cold Start. }\label{alg:prm_active_learning}
				\begin{algorithmic}[1]
                        \State \textcolor{darkred}{// The difference with full data tuning is colored}.
					\Require Ensemble PRM \(P_\theta\), dataset \(\mathcal{D}=\{(q, s)\}\), uncertainty thresholds \(\delta_{pred}\)
					and \(\delta_{std}\), generative LLM \(M\), batch size \(B\), learning rate \(\eta\)
					\For{\(\mathcal{B} \subset \mathcal{D}\)}
					\State \(P_\theta(\mathcal{B}) \gets \text{Forward}(\mathcal{B})\)
					\State \(\widetilde{\mathcal{B}} = \{\}\)
                        \beginalgcolor{darkred}
					\For{\((q, s) \in \mathcal{B}\)}
					\If {\(\mathcal{U}^{\text{alea}}_\theta(s) \lor 
                    \mathcal{U}^{\text{epis}}_\theta(s)\)}
                        \Comment{Equation~\ref{equ:uncertainty_threshold}}
					\State \(\widetilde{\mathcal{B}} \gets \widetilde{\mathcal{B}} \cup \{(q, s)\}\)
					\EndIf
					\EndFor
                        \beginalgcolor{black}
					\State \(Y_{\widetilde{\mathcal{B}}} \gets \text{labeling}(\widetilde{\mathcal{B}})\) \Comment{Labeling using generative LLM}
					\State \(\mathcal{L} \gets
					\frac{1}{|\widetilde{\mathcal{B}}|}\sum_{(s, y)\in(\widetilde{\mathcal{B}},
						Y_{\widetilde{\mathcal{B}}})}\textcolor{darkred}{\mathcal{L}(s, y)}\)
					\textcolor{darkred}{\Comment{Equation~\ref{equ:loss}}}
					\State \(\nabla_{\theta} \mathcal{L} \gets \text{Backward}(\mathcal{L})\)
					\State \(\theta \gets \theta - \eta \nabla_{\theta} \mathcal{L}\)
					\EndFor
					\Ensure \(P_\theta\)
				\end{algorithmic}
			\end{algorithm}
		\end{minipage}
	}
\end{figure}

To mitigate this issue, we propose to leverage active learning to make the PRM proactively select the data that is most informative to train on.
Specifically, we train a PRM with ensemble heads to enable uncertainty estimation following~\citet{liang2022reward,gleave2022uncertainty}. As shown in Algorithm~\ref{alg:prm_active_learning},
We forward the data candidates to the ensemble PRM (\textcolor{darkred}{line 3}) and estimate the prediction uncertainty for each data point (\textcolor{darkred}{line 5-6}). 
Then we omit the data that the ensemble PRM is confident about (\textcolor{darkred}{line 7}) and only label the other retained data with a generative reasoning LLM (\textcolor{darkred}{line 10}). Then, we only backpropagate from the loss of labeled data (\textcolor{darkred}{line 11}).
By doing so, we could considerably reduce the labeling cost while maintaining the PRM performance. Now we introduce our two key differences with the original finetuning: \emph{ensemble PRM training} and \emph{uncertain data selection}.

\noindent
\textbf{Ensemble PRM Training.} In this work, we use ensemble of PRMs to estimate the epistemic uncertainty following~\citet{gleave2022uncertainty,liang2022reward}.
Specifically, we use a shared LLM backbone and build multiple binary classification heads on top of it. In our training,
the diversity of ensemble models is ensured by two ways: \((i)\) the random initialization of the head layers
and \((ii)\) a diversity regularization term~\citep{dwaracherla2024efficient}:
\(\mathcal{L}_{\mbox{div}} = \lambda \cdot \frac{1}{n}\sum_{i=1}^n ||\phi^i - \phi_{\text{init}}^i||_2,\)
where $\{\phi^i\}$ are the parameters of the ensemble heads and n is the number of ensemble heads. It is a \(L2\) term penalizing the distance between the ensemble head parameters and their initial parameters. Our training objective for the ensemble PRM is therefore formulated as follows
\begin{equation}\label{equ:loss}
    \mathcal{L}(y, s) = \frac{1}{n} \sum_{i=1}^{n} \left(\mathcal{L}_{BCE}(y, s|\theta,\phi^i) + \lambda ||\phi^i - \phi_{\text{init}}^i||_2 \right),
\end{equation}
where \(\theta\) denotes the backbone parameters and \(\mathcal{L}_BCE\) is from Equation~\ref{equ:prm_loss}, that computes the loss for a certain head.

\noindent
\textbf{Uncertain Data Selection.} Considering a batch of data candidates \(\mathcal{D}=\{(q,s)\}\), we
first forward the data to the ensemble PRM \(P_\theta\) to get the ensemble predictions \(P_\theta(\mathcal{D}) \in
\mathbb{R}^{n\times |\mathcal{D}|\times |s|}\). for each data \((q,
s) \in \mathcal{D}\), we could determine the hard-value aleatoric and epistemic uncertainty with pre-set thresholds. Briefly, the aleatoric (or epistemic) uncertainty is defined as 1 if uncertainty occurs at any step prior to the first predicted error step; otherwise, it is 0. A formal definition is as follows:
\begin{equation}\label{equ:uncertainty_threshold}
	\begin{split}
		\mathcal{U}^{\text{alea}}_{\theta}(s)  = \bigvee_{i=0}^{\mathcal{E}(s)}\left(\max\left(\mu(P_\theta(s_i)),
			1-\mu(P_\theta(s_i))\right) <
		\delta_{pred}\right) \,\, ; \,\,
		\mathcal{U}^{\text{epis}}_{\theta}(s)   = \bigvee_{i=0}^{\mathcal{E}(s)}\left(\sigma(P_\theta(s_i)) > \delta_{std}\right),
	\end{split}
\end{equation}
where \(\mu(\cdot)\) and \(\sigma(\cdot)\) are the mean and standard deviation of the ensemble predictions among
ensemble heads and \(\lor\) denotes the logical `OR' operation. 
Moreover, the \(\mathcal{E}(s)\) denotes the first
error step in the solution trajectory \(s\), defined as \(\mathcal{E}(s) = \min\{j \mid \mu(s_j) < \delta\}\), where
\(\delta\) is the threshold for the correctness, typically set to \(0.5\). This is because we only care about the
correctness of the steps before the first error step since it is genuinely difficult to define the correctness of the steps afterwards. 
By following the uncertainty estimation strategy, we retain the data in \(\mathcal{D}\) that satisfies
either \(\mathcal{U}^{\text{alea}}_{\theta}\) or \(\mathcal{U}^{\text{epis}}_{\theta}\) as \(\widetilde{\mathcal{D}}\).
Then we could leverage
expensive generative LLMs as judger~\citep{zheng2024processbench} to label the retained data in
\(\widetilde{\mathcal{D}}\).

\vspace{-0.2cm}
\section{Experiments}
\vspace{-0.2cm}

In Section~\ref{sec:exp_pool_based}, we first validate \ourmethod in a pool-based active learning setting using 100K labeled samples, including ablation studies on our uncertainty estimation strategy. 
Based on the optimal configuration, we then scale up to 1M unlabeled samples in Section~\ref{sec:exp_sota}, further proving our pipeline’s efficiency and effectiveness.

\vspace{-0.2cm}

\subsection{Pool-Based Active Learning}~\label{sec:exp_pool_based}
\vspace{-0.6cm}

\subsubsection{Experimental Settings}
To evaluate our active learning strategy's effectiveness, we first conduct experiments in an offline setting where \ourmethod iteratively selects the most informative examples from a large unlabeled pool as detailed in Algorithm~\ref{alg:prm_active_learning}. 
We establish a strong baseline by comparing against full data tuning, where a model is trained on the complete dataset labeled by a single annotator.
It is worth noting that as our data is randomly shuffled, the performance of full data tuning at intermediate training steps is essentially equivalent to the performance of random selection with the corresponding budget.

\noindent
\textbf{Evaluation Benchmark.} We utilize ProcessBench~\citep{zheng2024processbench} to evaluate the effectiveness of PRMs. The test data in ProcessBench contains intermediate step errors and requires the PRM to identify the first error step. ProcessBench contains four subsets, and we report the average F1 score following the original work.

\noindent
\textbf{Models.} We train \ourmethod based on Qwen2.5-Math-7B-Instruct. 

\noindent
\textbf{Training Dataset.} For dataset construction, we randomly select 100K data from Numinamath~\citep{numinamath} dataset after decontamination against the ProcessBench~\citep{zheng2024processbench} and PRMBench~\citep{song2025prmbench}.
We leverage Qwen-2.5-Math-7B-Instruct to generate CoT reasoning trajectories for the selected data and further use QwQ-32B as a judge to annotate the process correctness for all trajectories following \citet{zhang2025lessons}. 
For completeness, we provide the prompt template in Appendix~\ref{sec:judge_prompt}.

\vspace{-0.2cm}

\subsubsection{Experimental Results}

\begin{figure}
    \centering
    \vspace{-0.8cm}
    \sbox0{
    \subfigure(a){\includegraphics[width=0.45\linewidth]{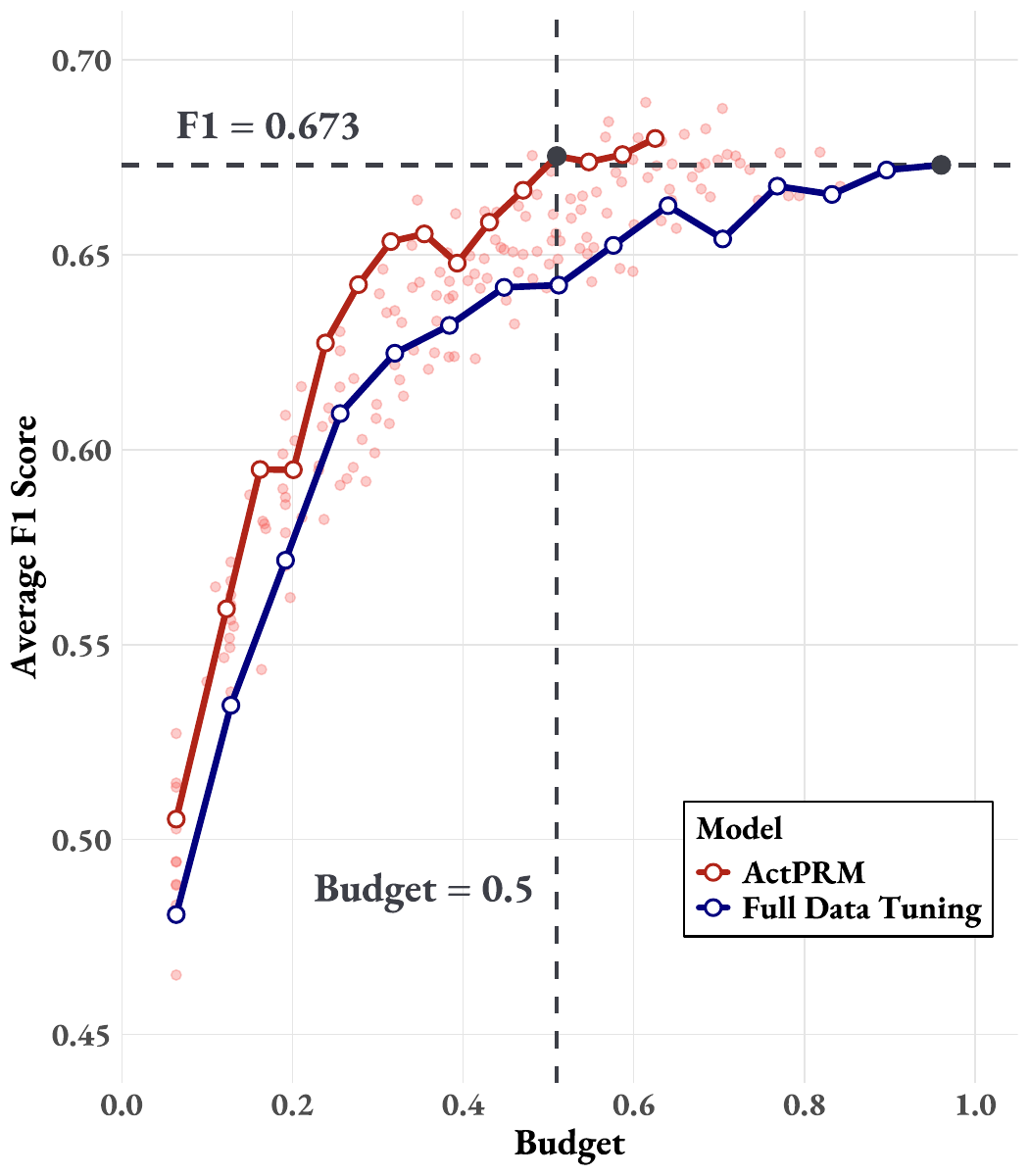}}}
    \usebox0\hfill\begin{minipage}[b][\ht0][s]{\wd0}
    \subfigure(b){\includegraphics[width=0.9\linewidth]{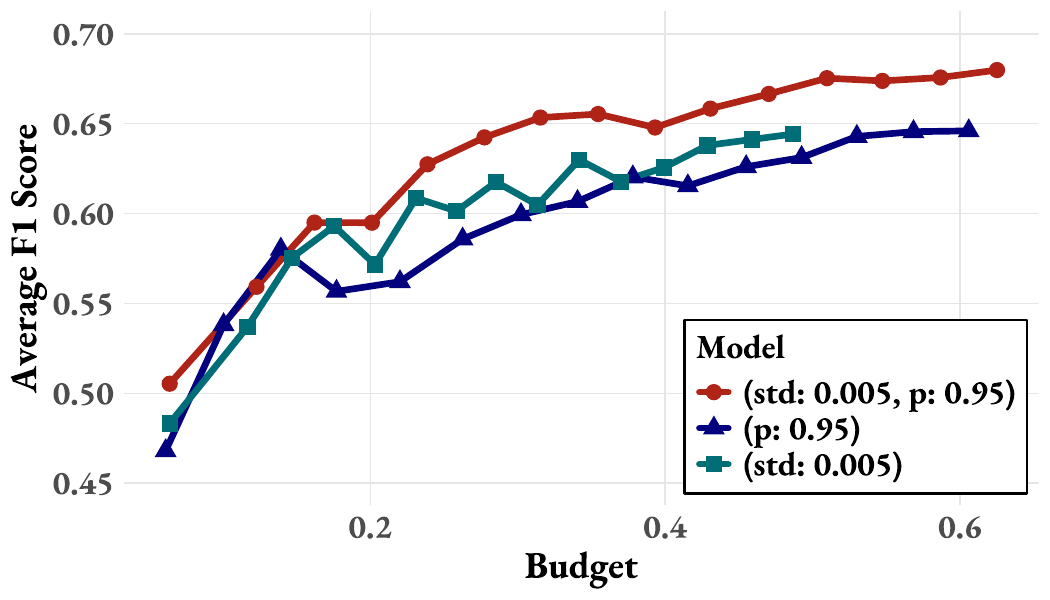}}
    \vfill
    \subfigure(c){\includegraphics[width=0.9\linewidth]{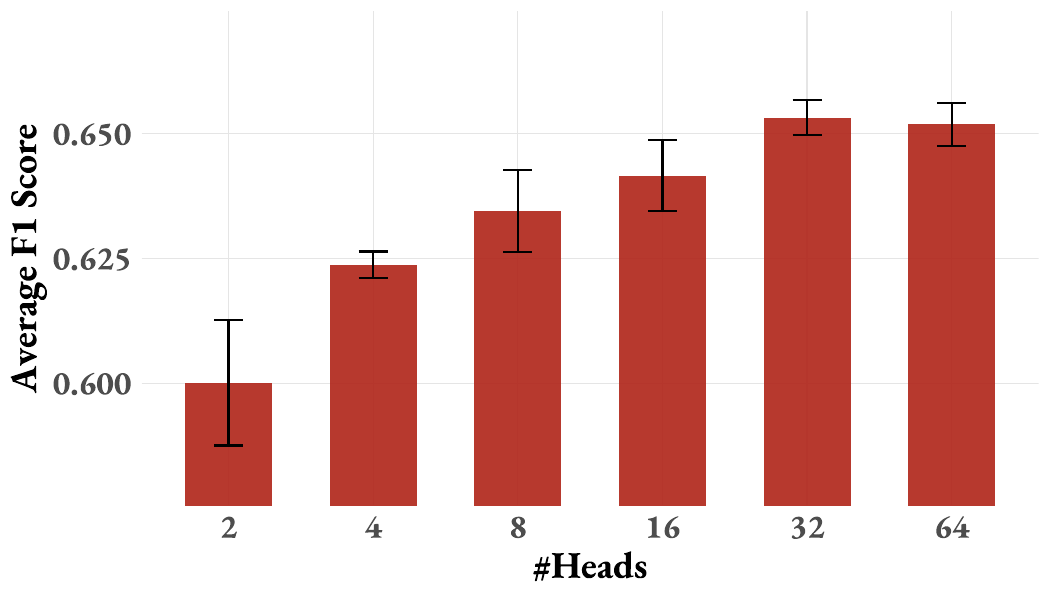}}
    \end{minipage}
    \caption{(a) Comparison of the average F1 score on ProcessBench between \ourmethod and random selection, plotted against the normalized budget positively correlated the number of labeled data instances consumed. The \emph{semi-transparent} points represent all results in grid searching w.r.t. \(\delta_{pred}\) and \(\delta_{std}\). For the highlighted \ourmethod curve in the figure, \(\delta_{pred}=0.95\) and \(\delta_{std}=0.005\).
    (b) Ablation: uncertainty estimation strategies.
    (c) Ablation: number of ensemble PRM heads.}
    \label{fig:pool_based_exp}
\end{figure}

\noindent
\textbf{\ourmethod achieves comparable performance while reducing annotation costs by 50\%.} We compare \ourmethod with full data tuning across a normalized budget, as illustrated in Figure~\ref{fig:pool_based_exp} (a). The results demonstrate that \ourmethod achieves an average F1 score of \(0.673\) on ProcessBench, matching baseline performance while using only half the annotation budget. Furthermore, \ourmethod consistently outperforms random selection under the same budget constraints. Notably, at 50\% budget, \ourmethod surpasses random selection by a significant margin of 3.3\%. at the end of pool-based active training, \ourmethod achieves a better performance of 0.680 on ProcessBench while consuming solely 62.5\% budget.

\noindent
\textbf{\ourmethod Consistently Outperforms Random Selection Under Diverse \(\delta_{pred}\) and \(\delta_{std}\).} As shown in Figure~\ref{fig:pool_based_exp} (a), the semi-transparent blue points represent all results of a grid searching over \(\delta_{pred} \in \{0.9, 0.95, 0.97\}\) and \(\delta_{std} \in \{0.01, 0.005, 0.002, 0.001\}\). One can see that most blue points are above the baseline (gray line) with the same budget, further demonstrating the effectiveness and robustness of \ourmethod.

\noindent
\textbf{Ablation Study on Uncertainty Estimation Strategies.}
We conduct an ablation study on uncertainty estimation strategies, i.e. using epistemic and aleatoric uncertainty. We selected the best setting (\(\delta_{std}=0.005, \delta_{pred}=0.95\)) searched by a grid search as in Figure~\ref{fig:pool_based_exp} and ablates epistemic and aleatoric uncertainty by setting \(\delta_{std}=\inf\) and \(\delta_{pred}=0.5\), respectively. As shown illustrated in Figure~\ref{fig:pool_based_exp} (b), solely use either epistemic or aleatoric uncertainty under-performs using both, indicating that epistemic and aleatoric uncertainty are complementary to each other.

\noindent
\textbf{Ablation Study on Number of Heads for Ensemble PRM.}
The number of heads for ensemble PRM controls how accurate our estimated epistemic uncertainty is. To find the trade-off between good estimation and computational overhead, we conduct an ablation study regarding it and show the results in Figure~\ref{fig:pool_based_exp} (c), where we only consider epistemic uncertainty by setting \(\delta_{std}=0.005, \delta_{pred}=0.5\) and report the averaged results with 3 runs. We empirically find that the performance continually grows with the number of heads and converges at about 32.

\vspace{-0.2cm}

\subsection{Achieving New SOTA Performance on ProcessBench (75.0\%) with Solely 6\% Annotation Cost.}~\label{sec:exp_sota}
\vspace{-0.3cm}

\begin{wrapfigure}{rt}{0.45\linewidth}
    \vspace{-10pt}
    \centering
    \includegraphics[width=\linewidth]{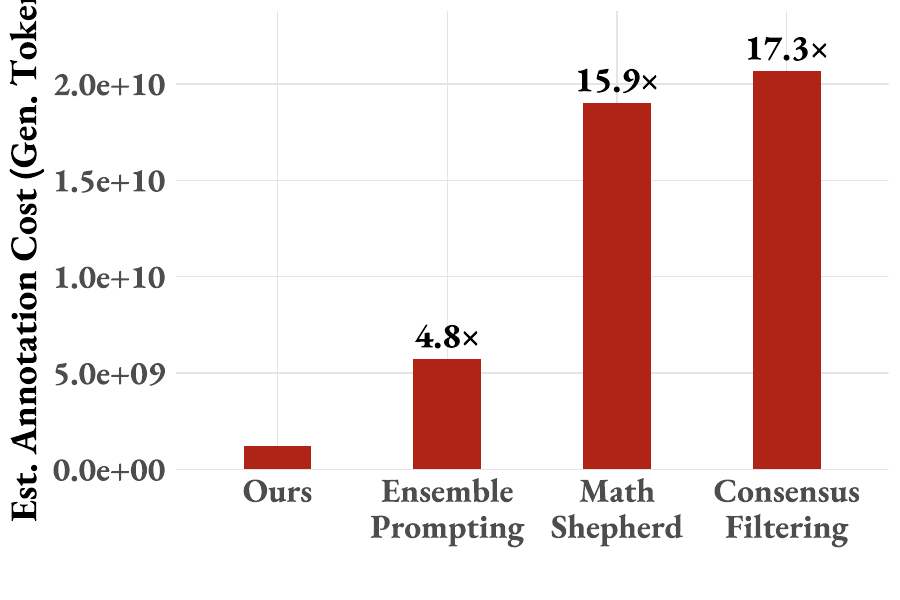}
    \vspace{-20pt}
    \caption{Estimated annotation costs (generated tokens) comparison between \ourmethod and popular methods, including Ensemble Prompting~\citep{tan2025universalprm}, MathShepherd~\citep{wang2024mathshepherd} and Consensus Filtering~\citep{zhang2025lessons}.}
    \label{fig:labeling_cost}
    \vspace{-20pt}
\end{wrapfigure}

 Obtaining high-quality process supervision labels is costly. To demonstrate the efficiency of \ourmethod, we evaluate it in a one-shot active learning setting. Starting with the model trained in Section~\ref{sec:exp_pool_based}, we select the most uncertain samples from over 1M+ unlabeled examples and annotate them using a powerful reasoning model.
 
 Figure~\ref{fig:labeling_cost} compares our estimated labeling costs with those of other real-world datasets for training PRMs, including MathShepherd~\citep{wang2024mathshepherd}, Consensus Filtering~\citep{zhang2025lessons}, and Ensemble Prompting~\citep{tan2025universalprm}. Since the training data for Consensus Filtering is not publicly available, we estimate costs based on our data statistics. We introduce our estimation strategy in Appendix~\ref{sec:labeling_cost_estimation}.
 
 Training a Qwen2.5-Math-7B-Instruct on our data, Ensemble Prompting data, MathShepherd data, and Consensus Filtering data yields \ourmethod, UniversalPRM, Qwen2.5-Math-7B-Math-Shepherd, and Qwen2.5-Math-PRM-7B in Table~\ref{tab:sota_processbench}. We evaluated the performance of models trained on this labeled data on both ProcessBench and PRMBench benchmarks.

\subsubsection{Experimental Settings} 

\noindent
\textbf{Data Filtering with \ourmethod.}
We used Qwen2.5-Math-7B-Instruct and Qwen2.5-Math-72B-Instruct to collect over 1 million (1,061,763) Chain-of-Thought (COT) trajectories from the Numinamath problem set~\citep{numinamath}, after decontamination against the test benchmarks. \ourmethod was then applied to filter out high-confidence (\(\delta_{pred}>0.95\) and \(\delta_{std}<0.005\) following Section~\ref{sec:exp_pool_based}) data instances that were unnecessary for training, retaining the remaining data for labeling and training. This process resulted in a final dataset of 563,030 PRM data points labeled by QwQ-32B, reducing annotation costs by 47.0\%.

\noindent
\textbf{Models.} 
Obtaining the dataset, we continually train our \ourmethod in Section~\ref{sec:exp_pool_based} on our filtered dataset. In addition, we empirically find that our retrained data is generally useful to other PRMs. Specifically, we also continually train Qwen2.5-Math-PRM-7B (the previous SOTA model on ProcessBench) on our constructed data. The resultant model is named \oursotamodel\footnote{X stands for extended version.}.

\noindent
\textbf{Benchmarks.} We use ProcessBench~\citep{zheng2024processbench} and PRMBench~\citep{song2025prmbench} to evaluate the effectiveness of our trained model. Different from ProcessBench that collects intermediate errors from real-world generative models, PRMBench heuristically builds intermediate errors by manipulating correct steps.

\noindent
\textbf{Baselines.} We compare with the following PRMs: 
\ding{182} \emph{Qwen2.5-Math-PRM-7B}~\citep{zhang2025lessons}: This model uses consensus filtering for labeling. It labels 860K data twice using two methods (LLM-as-judge [Zheng et al., 2024] and Mathshepherd [Wang et al., 2024]) and filters out 40\% of the data where the labels disagree. 
\ding{183} \emph{Pure-PRM-7B}~\citep{cheng2025pure}: A Qwen2.5-Math-based PRM trained on PRM800K using a two-stage strategy: warming up the PRM head and then fine-tuning the entire model. 
\ding{184}: \emph{EurusPRM-Stage2}~\citep{cui2025eurus}: A PRM resulting from the Implicit PRM approach~\citep{yuan2024implicitprm}, which derives process rewards from an ORM. 
\ding{185} \emph{Universal-PRM}~\citep{tan2025universalprm}: A Qwen2.5-Math-based model trained with data augmentation techniques like ensemble prompting and reverse verification. 
\ding{186} \emph{Math-Shepherd-PRM-7B}~\citep{wang2024mathshepherd}: a PRM trained on process labels that estimates hard Q-values for the policy model. 
\ding{187} \emph{Qwen2.5-Math-7B-Math-Shepherd}~\citep{zhang2025lessons}: a PRM trained on 860K data labeled using MathShepherd.
\ding{188} \emph{Ensemble-PRM-PRM800K (ours)}: a model with ensemble heads trained by ourselves on PRM800K without active learning.

\vspace{-0.2cm}

\subsubsection{Experimental Results}
\begin{table}[t]
\centering
\vspace{-0.7cm}
\resizebox{\textwidth}{!}{
\begin{tabular}{lccccc}
\toprule
\textbf{Models} & \textbf{GSM8K} & \textbf{MATH} & \textbf{\makecell[c]{Olympiad\\Bench}} & \textbf{OmniMath} & \textbf{Average F1} \\
\midrule
\multicolumn{6}{l}{\textbf{\emph{LLM-as-judge}}} \\
o1-Mini\(^\diamond\)  & 0.932 & 0.889 & 0.872 & 0.824 & 0.879 \\
\makecell[l]{Deepseek-R1-Distill-32B} & 0.817 & 0.739 & 0.659 & 0.585 & 0.700 \\
QwQ-32B  & 0.871 & 0.834 & 0.787 & 0.771 & 0.816 \\
\midrule
\multicolumn{6}{l}{\textbf{\emph{Process Reward Models (72B)}}} \\
Qwen2.5-Math-PRM-72B\(^\diamond\)  & 0.873 & 0.806 & 0.743 & 0.711 & 0.783 \\
\midrule
\multicolumn{6}{l}{\textbf{\emph{Process Reward Models (7B+)}}} \\
Math-Shepherd-PRM-7B\(^\diamond\) & 0.479 & 0.295 & 0.248 & 0.238 & 0.315 \\
Qwen2.5-Math-7B-Math-Shepherd\(^\diamond\) & 0.625 & 0.316 & 0.137 & 0.077 & 0.289 \\
EurusPRM-Stage2\(^\diamond\) & 0.473 & 0.357 & 0.212 & 0.209 & 0.313 \\
Qwen2.5-Math-7B-PRM800K\(^\diamond\)  & 0.683 & 0.626 & 0.507 & 0.443 & 0.565 \\
\makecell[l]{Ensemble-PRM-PRM800K (ours)}  & 0.705 & 0.630 & 0.472 & 0.433 & 0.560 \\
PURE-PRM-7B  & 0.690 & 0.665 & 0.484 & 0.459 & 0.575 \\
Qwen2.5-Math-PRM-7B\(^\diamond\)  & 0.824 & 0.776 & 0.675 & 0.663 & 0.735 \\
Universal-PRM & \textbf{0.858} & 0.777 & 0.676 & 0.664 & 0.743 \\
\addlinespace[0.2em]
\hdashline
\addlinespace[0.3em]
\rowcolor{gray!20}
\ourmethod (ours)  & 0.816 & 0.798 & 0.714 & 0.670 & 0.750 \\
\rowcolor{gray!20}
\oursotamodel (ours) & 0.827 & \textbf{0.820} & \textbf{0.720}	& \textbf{0.673} & \textbf{0.760}\\
\bottomrule
\end{tabular}}
\caption{Performance comparison on ProcessBench. 
We report the results in the same calculation method with ProcessBench. \(^\diamond\) denotes the results are from Qwen PRM's report~\citep{zhang2025lessons}.
}
\label{tab:sota_processbench}
\end{table}


\begin{table}[t]
\centering
\vspace{-0.2cm}

\resizebox{0.9\textwidth}{!}{ 
\begin{tabular}{llcccc}
\toprule
\textbf{\#} & \textbf{Models} & \textbf{Simlicity} & \textbf{Soundness} & \textbf{Sensitivity} & \textbf{Average} \\
\midrule
& \multicolumn{5}{l}{\textbf{\emph{LLM-as-judge}}} \\
1 & Gemini-2.0-thinking-exp-1219 & 0.662 & 0.718 & 0.753 & 0.688 \\
1 & o1-mini & 0.646 & 0.721 & 0.755 & 0.688 \\
4 & GPT-4o & 0.597 & 0.709 & 0.758 & 0.668 \\
6 & Gemini-2.0-flash-exp & 0.627 & 0.673 & 0.754 & 0.660 \\
\midrule
& \multicolumn{5}{l}{\textbf{\emph{Process Reward Models (72B)}}} \\
3 & Qwen-2.5-Math-PRM-72B & 0.546 & 0.739 & 0.770 & 0.682 \\
\midrule
& \multicolumn{5}{l}{\textbf{\emph{Process Reward Models (7B+)}}} \\
7 & Qwen2.5-Math-PRM-7B & 0.521 & 0.710 & 0.755 & 0.655 \\
9 & Pure-PRM-7B & 0.522 & 0.702 & \textbf{0.758} & 0.653 \\
\addlinespace[0.2em]
\hdashline
\addlinespace[0.3em]
\rowcolor{gray!20}
7 & \ourmethod (ours) & 0.536 & 0.713 & 0.752 & 0.655 \\
\rowcolor{gray!20} 
5 & \oursotamodel (ours) & \textbf{0.545} & \textbf{0.727} & 0.756 & \textbf{0.667}\\
\bottomrule
\end{tabular}
}
\caption{Performance comparison on PRMBench. 
All results of the other models are from the official leaderboard. \textbf{\#} denotes the ranking.
}
\label{tab:sota_prmbench}
\end{table}

\noindent
\textbf{\ourmethod and \oursotamodel achieve new SOTA performance on ProcessBench compared with same size models.} The evaluation results on ProcessBench are shown in Table~\ref{tab:sota_processbench}. \ourmethod achieves an average F1 score of \(0.750\), outperforming Qwen2.5-Math-PRM-7B by a margin of 1.5\%. Furthermore, \oursotamodel training based on Qwen2.5-Math-PRM-7B achieves a new SOTA performance on ProcessBench with an average F1 of 0.760, outperforming the second-place model (Universal-PRM) with a margin of 1.7\% and improve the performance of Qwen2.5-Math-PRM-7B by a significant margin of 2.5\%.

\noindent
\textbf{QwQ-32B (our PRM label annotator) outperforms all PRMs on ProcessBench.} As shown in Table~\ref{tab:sota_processbench}, QwQ-32B outperforms all PRMs including 72B models. It indicates the reliability of utilizing QwQ-32B as a PRM label annotator as it provides a high empirical upperbound for the training PRMs.

\noindent
\textbf{\oursotamodel achieves new SOTA performance on PRMBench, on-par with GPT-4o.} We further test our models on PRMBench and show the results in Table~\ref{tab:sota_prmbench}. 
As on the leaderboard, \ourmethod achieves the best performance within 7B PRMs and \oursotamodel achieves new SOTA performance (0.667), outperforming the other models by a large margin of at least \(1.2\%\) and on-par with GPT-4o~\citep{openai2024gpt4ocard}.

\begin{figure}[t]
\centering
\vspace{-0.7cm}

\includegraphics[width=0.9\linewidth]{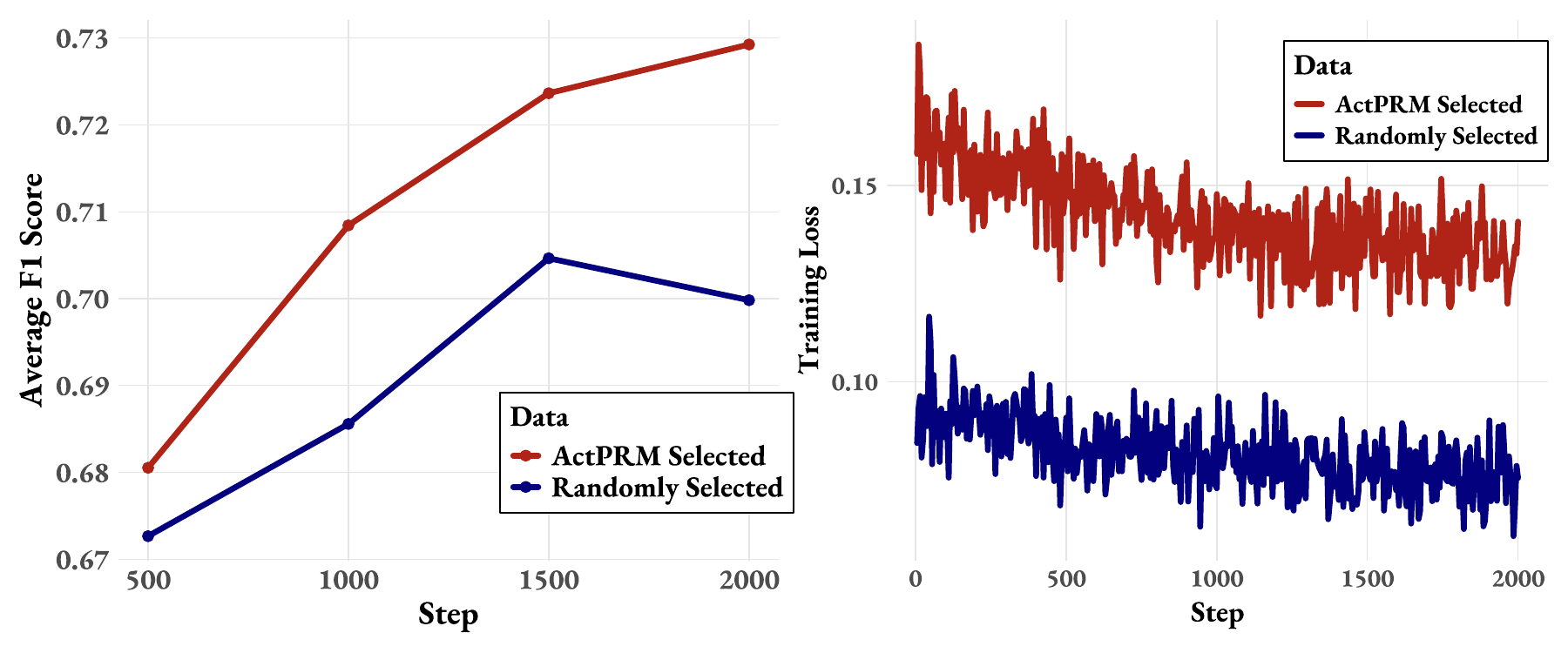}
\vspace{-0.3cm}

\caption{ProcessBench performance (\emph{left}) and training loss (\emph{right}): ActPRM v.s. random data selection on 1M NuminaMath Rollouts.} 
\label{fig:ablation_random_selection}
\end{figure}

\vspace{-0.2cm}

\subsubsection{Comparative Experiment with Random Selection}

A potential concern is that while \ourmethod achieves state-of-the-art (SOTA) performance on several benchmarks, this success might be attributed solely to the high quality of our collected data pool, rather than the method itself. To address this, we conducted a comparative study with random selection. Specifically, we randomly selected 256K data points from our retained dataset as the experimental group. For the control group, we randomly selected the same number of data points from the entire data pool (over 1M) and used the same annotator to label any unlabeled data (i.e., data not in the retained set). We then continually train \ourmethod checkpoint, as in Sec~\ref{sec:exp_pool_based}, on both datasets. The results, including performance on ProcessBench and training loss, are shown in Figure~\ref{fig:ablation_random_selection}.

\noindent
\textbf{\ourmethod outperforms random selection of the same amount of data.} 
As illustrated in Figure~\ref{fig:ablation_random_selection} (left), the model trained on data selected by \ourmethod consistently achieves significantly better results than the model trained on randomly selected data. To further validate this, we compare their training losses in Figure~\ref{fig:ablation_random_selection} (right). The model trained on \ourmethod-selected data exhibits a consistently higher training loss, with a margin of 0.05, suggesting that the data selected by \ourmethod is more challenging and informative, thereby enhancing the learning process.
\vspace{-0.2cm}

\section{Conclusion and Future Work}\label{sec:conclusion}
\vspace{-0.2cm}

In this work, we address the high annotation costs associated with training Process Reward Models (PRMs) by proposing \ourmethod, an uncertainty-aware active learning framework that selectively annotates the most informative reasoning steps. By leveraging an ensemble PRM to estimate uncertainty and strategically labeling only uncertain data, \ourmethod significantly reduces annotation costs while maintaining competitive performance. Extensive experiments demonstrate that \ourmethod achieves a new state-of-the-art (75.0\% on ProcessBench) with merely at most 20\% of the labeling budget required by prior methods. Our results highlight the potential of efficient data selection for scalable PRM training, and we commit to releasing all models, datasets, and code to foster further research in this direction.

To further enhance PRM's performance, several promising directions can be explored. First, leveraging larger base models and more advanced LLM judges (e.g., O1-mini) could yield significant improvements. Second, implementing the framework in an online setting would ultimately enable PRM to iteratively refine its performance through active learning. Additionally, integrating online PRM training with reinforcement learning frameworks—such as actor-critic methods—presents an exciting avenue for research.

\bibliography{colm2025_conference}

\begin{thebibliography}{27}
\providecommand{\natexlab}[1]{#1}
\providecommand{\url}[1]{\texttt{#1}}
\expandafter\ifx\csname urlstyle\endcsname\relax
  \providecommand{\doi}[1]{doi: #1}\else
  \providecommand{\doi}{doi: \begingroup \urlstyle{rm}\Url}\fi

\bibitem[Amodei et~al.(2016)Amodei, Olah, Steinhardt, Christiano, Schulman, and Mané]{amodei2016concreteproblemsaisafety}
Dario Amodei, Chris Olah, Jacob Steinhardt, Paul Christiano, John Schulman, and Dan Mané.
\newblock Concrete problems in ai safety, 2016.
\newblock URL \url{https://arxiv.org/abs/1606.06565}.

\bibitem[Cheng et~al.(2025)Cheng, Li, Xiong, Shao, and Lv]{cheng2025pure}
Jie Cheng, Lijun Li, Gang Xiong, Jing Shao, and Yisheng Lv.
\newblock Stop gamma decay: Min-form credit assignment is all process reward model needs for reasoning, 2025.
\newblock Notion Blog.

\bibitem[Coste et~al.(2024)Coste, Anwar, Kirk, and Krueger]{coste2024reward}
Thomas Coste, Usman Anwar, Robert Kirk, and David Krueger.
\newblock Reward {{Model Ensembles Help Mitigate Overoptimization}}, 2024.

\bibitem[Cui et~al.(2025)Cui, Yuan, Wang, Wang, Li, He, Fan, Yu, Xu, Chen, et~al.]{cui2025eurus}
Ganqu Cui, Lifan Yuan, Zefan Wang, Hanbin Wang, Wendi Li, Bingxiang He, Yuchen Fan, Tianyu Yu, Qixin Xu, Weize Chen, et~al.
\newblock Process reinforcement through implicit rewards.
\newblock \emph{arXiv preprint arXiv:2502.01456}, 2025.

\bibitem[DeepSeek-AI et~al.(2025)DeepSeek-AI, Guo, Yang, Zhang, Song, Zhang, Xu, Zhu, Ma, Wang, Bi, Zhang, Yu, Wu, Wu, Gou, Shao, Li, Gao, Liu, Xue, Wang, Wu, Feng, Lu, Zhao, Deng, Zhang, Ruan, Dai, Chen, Ji, Li, Lin, Dai, Luo, Hao, Chen, Li, Zhang, Bao, Xu, Wang, Ding, Xin, Gao, Qu, Li, Guo, Li, Wang, Chen, Yuan, Qiu, Li, Cai, Ni, Liang, Chen, Dong, Hu, Gao, Guan, Huang, Yu, Wang, Zhang, Zhao, Wang, Zhang, Xu, Xia, Zhang, Zhang, Tang, Li, Wang, Li, Tian, Huang, Zhang, Wang, Chen, Du, Ge, Zhang, Pan, Wang, Chen, Jin, Chen, Lu, Zhou, Chen, Ye, Wang, Yu, Zhou, Pan, Li, Zhou, Wu, Ye, Yun, Pei, Sun, Wang, Zeng, Zhao, Liu, Liang, Gao, Yu, Zhang, Xiao, An, Liu, Wang, Chen, Nie, Cheng, Liu, Xie, Liu, Yang, Li, Su, Lin, Li, Jin, Shen, Chen, Sun, Wang, Song, Zhou, Wang, Shan, Li, Wang, Wei, Zhang, Xu, Li, Zhao, Sun, Wang, Yu, Zhang, Shi, Xiong, He, Piao, Wang, Tan, Ma, Liu, Guo, Ou, Wang, Gong, Zou, He, Xiong, Luo, You, Liu, Zhou, Zhu, Xu, Huang, Li, Zheng, Zhu, Ma, Tang, Zha, Yan, Ren, Ren, Sha, Fu, Xu, Xie, Zhang,
  Hao, Ma, Yan, Wu, Gu, Zhu, Liu, Li, Xie, Song, Pan, Huang, Xu, Zhang, and Zhang]{deepseekr1}
DeepSeek-AI, Daya Guo, Dejian Yang, Haowei Zhang, Junxiao Song, Ruoyu Zhang, Runxin Xu, Qihao Zhu, Shirong Ma, Peiyi Wang, Xiao Bi, Xiaokang Zhang, Xingkai Yu, Yu~Wu, Z.~F. Wu, Zhibin Gou, Zhihong Shao, Zhuoshu Li, Ziyi Gao, Aixin Liu, Bing Xue, Bingxuan Wang, Bochao Wu, Bei Feng, Chengda Lu, Chenggang Zhao, Chengqi Deng, Chenyu Zhang, Chong Ruan, Damai Dai, Deli Chen, Dongjie Ji, Erhang Li, Fangyun Lin, Fucong Dai, Fuli Luo, Guangbo Hao, Guanting Chen, Guowei Li, H.~Zhang, Han Bao, Hanwei Xu, Haocheng Wang, Honghui Ding, Huajian Xin, Huazuo Gao, Hui Qu, Hui Li, Jianzhong Guo, Jiashi Li, Jiawei Wang, Jingchang Chen, Jingyang Yuan, Junjie Qiu, Junlong Li, J.~L. Cai, Jiaqi Ni, Jian Liang, Jin Chen, Kai Dong, Kai Hu, Kaige Gao, Kang Guan, Kexin Huang, Kuai Yu, Lean Wang, Lecong Zhang, Liang Zhao, Litong Wang, Liyue Zhang, Lei Xu, Leyi Xia, Mingchuan Zhang, Minghua Zhang, Minghui Tang, Meng Li, Miaojun Wang, Mingming Li, Ning Tian, Panpan Huang, Peng Zhang, Qiancheng Wang, Qinyu Chen, Qiushi Du, Ruiqi Ge, Ruisong
  Zhang, Ruizhe Pan, Runji Wang, R.~J. Chen, R.~L. Jin, Ruyi Chen, Shanghao Lu, Shangyan Zhou, Shanhuang Chen, Shengfeng Ye, Shiyu Wang, Shuiping Yu, Shunfeng Zhou, Shuting Pan, S.~S. Li, Shuang Zhou, Shaoqing Wu, Shengfeng Ye, Tao Yun, Tian Pei, Tianyu Sun, T.~Wang, Wangding Zeng, Wanjia Zhao, Wen Liu, Wenfeng Liang, Wenjun Gao, Wenqin Yu, Wentao Zhang, W.~L. Xiao, Wei An, Xiaodong Liu, Xiaohan Wang, Xiaokang Chen, Xiaotao Nie, Xin Cheng, Xin Liu, Xin Xie, Xingchao Liu, Xinyu Yang, Xinyuan Li, Xuecheng Su, Xuheng Lin, X.~Q. Li, Xiangyue Jin, Xiaojin Shen, Xiaosha Chen, Xiaowen Sun, Xiaoxiang Wang, Xinnan Song, Xinyi Zhou, Xianzu Wang, Xinxia Shan, Y.~K. Li, Y.~Q. Wang, Y.~X. Wei, Yang Zhang, Yanhong Xu, Yao Li, Yao Zhao, Yaofeng Sun, Yaohui Wang, Yi~Yu, Yichao Zhang, Yifan Shi, Yiliang Xiong, Ying He, Yishi Piao, Yisong Wang, Yixuan Tan, Yiyang Ma, Yiyuan Liu, Yongqiang Guo, Yuan Ou, Yuduan Wang, Yue Gong, Yuheng Zou, Yujia He, Yunfan Xiong, Yuxiang Luo, Yuxiang You, Yuxuan Liu, Yuyang Zhou, Y.~X. Zhu,
  Yanhong Xu, Yanping Huang, Yaohui Li, Yi~Zheng, Yuchen Zhu, Yunxian Ma, Ying Tang, Yukun Zha, Yuting Yan, Z.~Z. Ren, Zehui Ren, Zhangli Sha, Zhe Fu, Zhean Xu, Zhenda Xie, Zhengyan Zhang, Zhewen Hao, Zhicheng Ma, Zhigang Yan, Zhiyu Wu, Zihui Gu, Zijia Zhu, Zijun Liu, Zilin Li, Ziwei Xie, Ziyang Song, Zizheng Pan, Zhen Huang, Zhipeng Xu, Zhongyu Zhang, and Zhen Zhang.
\newblock Deepseek-r1: Incentivizing reasoning capability in llms via reinforcement learning, 2025.
\newblock URL \url{https://arxiv.org/abs/2501.12948}.

\bibitem[Dwaracherla et~al.(2024)Dwaracherla, Asghari, Hao, and Van~Roy]{dwaracherla2024efficient}
Vikranth Dwaracherla, Seyed~Mohammad Asghari, Botao Hao, and Benjamin Van~Roy.
\newblock Efficient exploration for llms.
\newblock In \emph{International Conference on Machine Learning}, 2024.

\bibitem[Gleave \& Irving(2022)Gleave and Irving]{gleave2022uncertainty}
Adam Gleave and Geoffrey Irving.
\newblock Uncertainty {{Estimation}} for {{Language Reward Models}}, 2022.

\bibitem[Hendrycks et~al.(2021)Hendrycks, Burns, Kadavath, Arora, Basart, Tang, Song, and Steinhardt]{hendrycks2021math}
Dan Hendrycks, Collin Burns, Saurav Kadavath, Akul Arora, Steven Basart, Eric Tang, Dawn Song, and Jacob Steinhardt.
\newblock Measuring mathematical problem solving with the math dataset, 2021.
\newblock URL \url{https://arxiv.org/abs/2103.03874}.

\bibitem[Li et~al.(2024)Li, Beeching, Tunstall, Lipkin, Soletskyi, Huang, Rasul, Yu, Jiang, Shen, Qin, Dong, Zhou, Fleureau, Lample, and Polu]{numinamath}
Jia Li, Edward Beeching, Lewis Tunstall, Ben Lipkin, Roman Soletskyi, Shengyi~Costa Huang, Kashif Rasul, Longhui Yu, Albert Jiang, Ziju Shen, Zihan Qin, Bin Dong, Li~Zhou, Yann Fleureau, Guillaume Lample, and Stanislas Polu.
\newblock Numinamath.
\newblock \url{ [https://huggingface.co/AI-MO/NuminaMath-1.5](https://github.com/project-numina/aimo-progress-prize/blob/main/report/numina_dataset.pdf) }, 2024.

\bibitem[Li \& Li(2024)Li and Li]{li2024qvalueranking}
Wendi Li and Yixuan Li.
\newblock Process reward model with q-value rankings.
\newblock \emph{arXiv preprint arXiv:2410.11287}, 2024.

\bibitem[Liang et~al.(2022)Liang, Shu, Lee, and Abbeel]{liang2022reward}
Xinran Liang, Katherine Shu, Kimin Lee, and Pieter Abbeel.
\newblock Reward {{Uncertainty}} for {{Exploration}} in {{Preference-based Reinforcement Learning}}, 2022.

\bibitem[Lightman et~al.(2023)Lightman, Kosaraju, Burda, Edwards, Baker, Lee, Leike, Schulman, Sutskever, and Cobbe]{lightman2023lets}
Hunter Lightman, Vineet Kosaraju, Yura Burda, Harri Edwards, Bowen Baker, Teddy Lee, Jan Leike, John Schulman, Ilya Sutskever, and Karl Cobbe.
\newblock Let's {{Verify Step}} by {{Step}}, 2023.

\bibitem[Liu et~al.(2024)Liu, Chen, Du, Lee, and Lin]{liu2024sampleefficientalignmentllms}
Zichen Liu, Changyu Chen, Chao Du, Wee~Sun Lee, and Min Lin.
\newblock Sample-efficient alignment for llms, 2024.
\newblock URL \url{https://arxiv.org/abs/2411.01493}.

\bibitem[Luo et~al.(2024)Luo, Liu, Liu, Phatale, Guo, Lara, Li, Shu, Zhu, Meng, Sun, and Rastogi]{luo2024improve}
Liangchen Luo, Yinxiao Liu, Rosanne Liu, Samrat Phatale, Meiqi Guo, Harsh Lara, Yunxuan Li, Lei Shu, Yun Zhu, Lei Meng, Jiao Sun, and Abhinav Rastogi.
\newblock Improve {{Mathematical Reasoning}} in {{Language Models}} by {{ Automated Process Supervision}}, 2024.

\bibitem[Mehta et~al.(2023)Mehta, Das, Neopane, Dai, Bogunovic, Schneider, and Neiswanger]{mehta2023sample}
Viraj Mehta, Vikramjeet Das, Ojash Neopane, Yijia Dai, Ilija Bogunovic, Jeff Schneider, and Willie Neiswanger.
\newblock Sample efficient reinforcement learning from human feedback via active exploration.
\newblock \emph{arXiv preprint arXiv:2312.00267}, 2023.

\bibitem[Melo et~al.(2024)Melo, Tigas, Abate, and Gal]{melo2024deepbayesianactivelearning}
Luckeciano~C. Melo, Panagiotis Tigas, Alessandro Abate, and Yarin Gal.
\newblock Deep bayesian active learning for preference modeling in large language models, 2024.
\newblock URL \url{https://arxiv.org/abs/2406.10023}.

\bibitem[OpenAI et~al.(2024{\natexlab{a}})OpenAI, :, Hurst, Lerer, Goucher, Perelman, Ramesh, Clark, Ostrow, Welihinda, Hayes, Radford, Madry, Baker-Whitcomb, Beutel, Borzunov, Carney, Chow, Kirillov, Nichol, Paino, Renzin, Passos, Kirillov, Christakis, Conneau, Kamali, Jabri, Moyer, Tam, Crookes, Tootoochian, Tootoonchian, Kumar, Vallone, Karpathy, Braunstein, Cann, Codispoti, Galu, Kondrich, Tulloch, Mishchenko, Baek, Jiang, Pelisse, Woodford, Gosalia, Dhar, Pantuliano, Nayak, Oliver, Zoph, Ghorbani, Leimberger, Rossen, Sokolowsky, Wang, Zweig, Hoover, Samic, McGrew, Spero, Giertler, Cheng, Lightcap, Walkin, Quinn, Guarraci, Hsu, Kellogg, Eastman, Lugaresi, Wainwright, Bassin, Hudson, Chu, Nelson, Li, Shern, Conger, Barette, Voss, Ding, Lu, Zhang, Beaumont, Hallacy, Koch, Gibson, Kim, Choi, McLeavey, Hesse, Fischer, Winter, Czarnecki, Jarvis, Wei, Koumouzelis, Sherburn, Kappler, Levin, Levy, Carr, Farhi, Mely, Robinson, Sasaki, Jin, Valladares, Tsipras, Li, Nguyen, Findlay, Oiwoh, Wong, Asdar, Proehl, Yang,
  Antonow, Kramer, Peterson, Sigler, Wallace, Brevdo, Mays, Khorasani, Such, Raso, Zhang, von Lohmann, Sulit, Goh, Oden, Salmon, Starace, Brockman, Salman, Bao, Hu, Wong, Wang, Schmidt, Whitney, Jun, Kirchner, de~Oliveira~Pinto, Ren, Chang, Chung, Kivlichan, O'Connell, O'Connell, Osband, Silber, Sohl, Okuyucu, Lan, Kostrikov, Sutskever, Kanitscheider, Gulrajani, Coxon, Menick, Pachocki, Aung, Betker, Crooks, Lennon, Kiros, Leike, Park, Kwon, Phang, Teplitz, Wei, Wolfe, Chen, Harris, Varavva, Lee, Shieh, Lin, Yu, Weng, Tang, Yu, Jang, Candela, Beutler, Landers, Parish, Heidecke, Schulman, Lachman, McKay, Uesato, Ward, Kim, Huizinga, Sitkin, Kraaijeveld, Gross, Kaplan, Snyder, Achiam, Jiao, Lee, Zhuang, Harriman, Fricke, Hayashi, Singhal, Shi, Karthik, Wood, Rimbach, Hsu, Nguyen, Gu-Lemberg, Button, Liu, Howe, Muthukumar, Luther, Ahmad, Kai, Itow, Workman, Pathak, Chen, Jing, Guy, Fedus, Zhou, Mamitsuka, Weng, McCallum, Held, Ouyang, Feuvrier, Zhang, Kondraciuk, Kaiser, Hewitt, Metz, Doshi, Aflak, Simens, Boyd,
  Thompson, Dukhan, Chen, Gray, Hudnall, Zhang, Aljubeh, Litwin, Zeng, Johnson, Shetty, Gupta, Shah, Yatbaz, Yang, Zhong, Glaese, Chen, Janner, Lampe, Petrov, Wu, Wang, Fradin, Pokrass, Castro, de~Castro, Pavlov, Brundage, Wang, Khan, Murati, Bavarian, Lin, Yesildal, Soto, Gimelshein, Cone, Staudacher, Summers, LaFontaine, Chowdhury, Ryder, Stathas, Turley, Tezak, Felix, Kudige, Keskar, Deutsch, Bundick, Puckett, Nachum, Okelola, Boiko, Murk, Jaffe, Watkins, Godement, Campbell-Moore, Chao, McMillan, Belov, Su, Bak, Bakkum, Deng, Dolan, Hoeschele, Welinder, Tillet, Pronin, Tillet, Dhariwal, Yuan, Dias, Lim, Arora, Troll, Lin, Lopes, Puri, Miyara, Leike, Gaubert, Zamani, Wang, Donnelly, Honsby, Smith, Sahai, Ramchandani, Huet, Carmichael, Zellers, Chen, Chen, Nigmatullin, Cheu, Jain, Altman, Schoenholz, Toizer, Miserendino, Agarwal, Culver, Ethersmith, Gray, Grove, Metzger, Hermani, Jain, Zhao, Wu, Jomoto, Wu, Shuaiqi, Xia, Phene, Papay, Narayanan, Coffey, Lee, Hall, Balaji, Broda, Stramer, Xu, Gogineni,
  Christianson, Sanders, Patwardhan, Cunninghman, Degry, Dimson, Raoux, Shadwell, Zheng, Underwood, Markov, Sherbakov, Rubin, Stasi, Kaftan, Heywood, Peterson, Walters, Eloundou, Qi, Moeller, Monaco, Kuo, Fomenko, Chang, Zheng, Zhou, Manassra, Sheu, Zaremba, Patil, Qian, Kim, Cheng, Zhang, He, Zhang, Jin, Dai, and Malkov]{openai2024gpt4ocard}
OpenAI, :, Aaron Hurst, Adam Lerer, Adam~P. Goucher, Adam Perelman, Aditya Ramesh, Aidan Clark, AJ~Ostrow, Akila Welihinda, Alan Hayes, Alec Radford, Aleksander Madry, Alex Baker-Whitcomb, Alex Beutel, Alex Borzunov, Alex Carney, Alex Chow, Alex Kirillov, Alex Nichol, Alex Paino, Alex Renzin, Alex~Tachard Passos, Alexander Kirillov, Alexi Christakis, Alexis Conneau, Ali Kamali, Allan Jabri, Allison Moyer, Allison Tam, Amadou Crookes, Amin Tootoochian, Amin Tootoonchian, Ananya Kumar, Andrea Vallone, Andrej Karpathy, Andrew Braunstein, Andrew Cann, Andrew Codispoti, Andrew Galu, Andrew Kondrich, Andrew Tulloch, Andrey Mishchenko, Angela Baek, Angela Jiang, Antoine Pelisse, Antonia Woodford, Anuj Gosalia, Arka Dhar, Ashley Pantuliano, Avi Nayak, Avital Oliver, Barret Zoph, Behrooz Ghorbani, Ben Leimberger, Ben Rossen, Ben Sokolowsky, Ben Wang, Benjamin Zweig, Beth Hoover, Blake Samic, Bob McGrew, Bobby Spero, Bogo Giertler, Bowen Cheng, Brad Lightcap, Brandon Walkin, Brendan Quinn, Brian Guarraci, Brian Hsu,
  Bright Kellogg, Brydon Eastman, Camillo Lugaresi, Carroll Wainwright, Cary Bassin, Cary Hudson, Casey Chu, Chad Nelson, Chak Li, Chan~Jun Shern, Channing Conger, Charlotte Barette, Chelsea Voss, Chen Ding, Cheng Lu, Chong Zhang, Chris Beaumont, Chris Hallacy, Chris Koch, Christian Gibson, Christina Kim, Christine Choi, Christine McLeavey, Christopher Hesse, Claudia Fischer, Clemens Winter, Coley Czarnecki, Colin Jarvis, Colin Wei, Constantin Koumouzelis, Dane Sherburn, Daniel Kappler, Daniel Levin, Daniel Levy, David Carr, David Farhi, David Mely, David Robinson, David Sasaki, Denny Jin, Dev Valladares, Dimitris Tsipras, Doug Li, Duc~Phong Nguyen, Duncan Findlay, Edede Oiwoh, Edmund Wong, Ehsan Asdar, Elizabeth Proehl, Elizabeth Yang, Eric Antonow, Eric Kramer, Eric Peterson, Eric Sigler, Eric Wallace, Eugene Brevdo, Evan Mays, Farzad Khorasani, Felipe~Petroski Such, Filippo Raso, Francis Zhang, Fred von Lohmann, Freddie Sulit, Gabriel Goh, Gene Oden, Geoff Salmon, Giulio Starace, Greg Brockman, Hadi
  Salman, Haiming Bao, Haitang Hu, Hannah Wong, Haoyu Wang, Heather Schmidt, Heather Whitney, Heewoo Jun, Hendrik Kirchner, Henrique~Ponde de~Oliveira~Pinto, Hongyu Ren, Huiwen Chang, Hyung~Won Chung, Ian Kivlichan, Ian O'Connell, Ian O'Connell, Ian Osband, Ian Silber, Ian Sohl, Ibrahim Okuyucu, Ikai Lan, Ilya Kostrikov, Ilya Sutskever, Ingmar Kanitscheider, Ishaan Gulrajani, Jacob Coxon, Jacob Menick, Jakub Pachocki, James Aung, James Betker, James Crooks, James Lennon, Jamie Kiros, Jan Leike, Jane Park, Jason Kwon, Jason Phang, Jason Teplitz, Jason Wei, Jason Wolfe, Jay Chen, Jeff Harris, Jenia Varavva, Jessica~Gan Lee, Jessica Shieh, Ji~Lin, Jiahui Yu, Jiayi Weng, Jie Tang, Jieqi Yu, Joanne Jang, Joaquin~Quinonero Candela, Joe Beutler, Joe Landers, Joel Parish, Johannes Heidecke, John Schulman, Jonathan Lachman, Jonathan McKay, Jonathan Uesato, Jonathan Ward, Jong~Wook Kim, Joost Huizinga, Jordan Sitkin, Jos Kraaijeveld, Josh Gross, Josh Kaplan, Josh Snyder, Joshua Achiam, Joy Jiao, Joyce Lee, Juntang
  Zhuang, Justyn Harriman, Kai Fricke, Kai Hayashi, Karan Singhal, Katy Shi, Kavin Karthik, Kayla Wood, Kendra Rimbach, Kenny Hsu, Kenny Nguyen, Keren Gu-Lemberg, Kevin Button, Kevin Liu, Kiel Howe, Krithika Muthukumar, Kyle Luther, Lama Ahmad, Larry Kai, Lauren Itow, Lauren Workman, Leher Pathak, Leo Chen, Li~Jing, Lia Guy, Liam Fedus, Liang Zhou, Lien Mamitsuka, Lilian Weng, Lindsay McCallum, Lindsey Held, Long Ouyang, Louis Feuvrier, Lu~Zhang, Lukas Kondraciuk, Lukasz Kaiser, Luke Hewitt, Luke Metz, Lyric Doshi, Mada Aflak, Maddie Simens, Madelaine Boyd, Madeleine Thompson, Marat Dukhan, Mark Chen, Mark Gray, Mark Hudnall, Marvin Zhang, Marwan Aljubeh, Mateusz Litwin, Matthew Zeng, Max Johnson, Maya Shetty, Mayank Gupta, Meghan Shah, Mehmet Yatbaz, Meng~Jia Yang, Mengchao Zhong, Mia Glaese, Mianna Chen, Michael Janner, Michael Lampe, Michael Petrov, Michael Wu, Michele Wang, Michelle Fradin, Michelle Pokrass, Miguel Castro, Miguel Oom~Temudo de~Castro, Mikhail Pavlov, Miles Brundage, Miles Wang, Minal
  Khan, Mira Murati, Mo~Bavarian, Molly Lin, Murat Yesildal, Nacho Soto, Natalia Gimelshein, Natalie Cone, Natalie Staudacher, Natalie Summers, Natan LaFontaine, Neil Chowdhury, Nick Ryder, Nick Stathas, Nick Turley, Nik Tezak, Niko Felix, Nithanth Kudige, Nitish Keskar, Noah Deutsch, Noel Bundick, Nora Puckett, Ofir Nachum, Ola Okelola, Oleg Boiko, Oleg Murk, Oliver Jaffe, Olivia Watkins, Olivier Godement, Owen Campbell-Moore, Patrick Chao, Paul McMillan, Pavel Belov, Peng Su, Peter Bak, Peter Bakkum, Peter Deng, Peter Dolan, Peter Hoeschele, Peter Welinder, Phil Tillet, Philip Pronin, Philippe Tillet, Prafulla Dhariwal, Qiming Yuan, Rachel Dias, Rachel Lim, Rahul Arora, Rajan Troll, Randall Lin, Rapha~Gontijo Lopes, Raul Puri, Reah Miyara, Reimar Leike, Renaud Gaubert, Reza Zamani, Ricky Wang, Rob Donnelly, Rob Honsby, Rocky Smith, Rohan Sahai, Rohit Ramchandani, Romain Huet, Rory Carmichael, Rowan Zellers, Roy Chen, Ruby Chen, Ruslan Nigmatullin, Ryan Cheu, Saachi Jain, Sam Altman, Sam Schoenholz, Sam
  Toizer, Samuel Miserendino, Sandhini Agarwal, Sara Culver, Scott Ethersmith, Scott Gray, Sean Grove, Sean Metzger, Shamez Hermani, Shantanu Jain, Shengjia Zhao, Sherwin Wu, Shino Jomoto, Shirong Wu, Shuaiqi, Xia, Sonia Phene, Spencer Papay, Srinivas Narayanan, Steve Coffey, Steve Lee, Stewart Hall, Suchir Balaji, Tal Broda, Tal Stramer, Tao Xu, Tarun Gogineni, Taya Christianson, Ted Sanders, Tejal Patwardhan, Thomas Cunninghman, Thomas Degry, Thomas Dimson, Thomas Raoux, Thomas Shadwell, Tianhao Zheng, Todd Underwood, Todor Markov, Toki Sherbakov, Tom Rubin, Tom Stasi, Tomer Kaftan, Tristan Heywood, Troy Peterson, Tyce Walters, Tyna Eloundou, Valerie Qi, Veit Moeller, Vinnie Monaco, Vishal Kuo, Vlad Fomenko, Wayne Chang, Weiyi Zheng, Wenda Zhou, Wesam Manassra, Will Sheu, Wojciech Zaremba, Yash Patil, Yilei Qian, Yongjik Kim, Youlong Cheng, Yu~Zhang, Yuchen He, Yuchen Zhang, Yujia Jin, Yunxing Dai, and Yury Malkov.
\newblock Gpt-4o system card, 2024{\natexlab{a}}.
\newblock URL \url{https://arxiv.org/abs/2410.21276}.

\bibitem[OpenAI et~al.(2024{\natexlab{b}})OpenAI, :, Jaech, Kalai, Lerer, Richardson, El-Kishky, Low, Helyar, Madry, Beutel, Carney, Iftimie, Karpenko, Passos, Neitz, Prokofiev, Wei, Tam, Bennett, Kumar, Saraiva, Vallone, Duberstein, Kondrich, Mishchenko, Applebaum, Jiang, Nair, Zoph, Ghorbani, Rossen, Sokolowsky, Barak, McGrew, Minaiev, Hao, Baker, Houghton, McKinzie, Eastman, Lugaresi, Bassin, Hudson, Li, de~Bourcy, Voss, Shen, Zhang, Koch, Orsinger, Hesse, Fischer, Chan, Roberts, Kappler, Levy, Selsam, Dohan, Farhi, Mely, Robinson, Tsipras, Li, Oprica, Freeman, Zhang, Wong, Proehl, Cheung, Mitchell, Wallace, Ritter, Mays, Wang, Such, Raso, Leoni, Tsimpourlas, Song, von Lohmann, Sulit, Salmon, Parascandolo, Chabot, Zhao, Brockman, Leclerc, Salman, Bao, Sheng, Andrin, Bagherinezhad, Ren, Lightman, Chung, Kivlichan, O'Connell, Osband, Gilaberte, Akkaya, Kostrikov, Sutskever, Kofman, Pachocki, Lennon, Wei, Harb, Twore, Feng, Yu, Weng, Tang, Yu, Candela, Palermo, Parish, Heidecke, Hallman, Rizzo, Gordon,
  Uesato, Ward, Huizinga, Wang, Chen, Xiao, Singhal, Nguyen, Cobbe, Shi, Wood, Rimbach, Gu-Lemberg, Liu, Lu, Stone, Yu, Ahmad, Yang, Liu, Maksin, Ho, Fedus, Weng, Li, McCallum, Held, Kuhn, Kondraciuk, Kaiser, Metz, Boyd, Trebacz, Joglekar, Chen, Tintor, Meyer, Jones, Kaufer, Schwarzer, Shah, Yatbaz, Guan, Xu, Yan, Glaese, Chen, Lampe, Malek, Wang, Fradin, McClay, Pavlov, Wang, Wang, Murati, Bavarian, Rohaninejad, McAleese, Chowdhury, Chowdhury, Ryder, Tezak, Brown, Nachum, Boiko, Murk, Watkins, Chao, Ashbourne, Izmailov, Zhokhov, Dias, Arora, Lin, Lopes, Gaon, Miyara, Leike, Hwang, Garg, Brown, James, Shu, Cheu, Greene, Jain, Altman, Toizer, Toyer, Miserendino, Agarwal, Hernandez, Baker, McKinney, Yan, Zhao, Hu, Santurkar, Chaudhuri, Zhang, Fu, Papay, Lin, Balaji, Sanjeev, Sidor, Broda, Clark, Wang, Gordon, Sanders, Patwardhan, Sottiaux, Degry, Dimson, Zheng, Garipov, Stasi, Bansal, Creech, Peterson, Eloundou, Qi, Kosaraju, Monaco, Pong, Fomenko, Zheng, Zhou, McCabe, Zaremba, Dubois, Lu, Chen, Cha, Bai, He,
  Zhang, Wang, Shao, and Li]{openai2024openaio1card}
OpenAI, :, Aaron Jaech, Adam Kalai, Adam Lerer, Adam Richardson, Ahmed El-Kishky, Aiden Low, Alec Helyar, Aleksander Madry, Alex Beutel, Alex Carney, Alex Iftimie, Alex Karpenko, Alex~Tachard Passos, Alexander Neitz, Alexander Prokofiev, Alexander Wei, Allison Tam, Ally Bennett, Ananya Kumar, Andre Saraiva, Andrea Vallone, Andrew Duberstein, Andrew Kondrich, Andrey Mishchenko, Andy Applebaum, Angela Jiang, Ashvin Nair, Barret Zoph, Behrooz Ghorbani, Ben Rossen, Benjamin Sokolowsky, Boaz Barak, Bob McGrew, Borys Minaiev, Botao Hao, Bowen Baker, Brandon Houghton, Brandon McKinzie, Brydon Eastman, Camillo Lugaresi, Cary Bassin, Cary Hudson, Chak~Ming Li, Charles de~Bourcy, Chelsea Voss, Chen Shen, Chong Zhang, Chris Koch, Chris Orsinger, Christopher Hesse, Claudia Fischer, Clive Chan, Dan Roberts, Daniel Kappler, Daniel Levy, Daniel Selsam, David Dohan, David Farhi, David Mely, David Robinson, Dimitris Tsipras, Doug Li, Dragos Oprica, Eben Freeman, Eddie Zhang, Edmund Wong, Elizabeth Proehl, Enoch Cheung, Eric
  Mitchell, Eric Wallace, Erik Ritter, Evan Mays, Fan Wang, Felipe~Petroski Such, Filippo Raso, Florencia Leoni, Foivos Tsimpourlas, Francis Song, Fred von Lohmann, Freddie Sulit, Geoff Salmon, Giambattista Parascandolo, Gildas Chabot, Grace Zhao, Greg Brockman, Guillaume Leclerc, Hadi Salman, Haiming Bao, Hao Sheng, Hart Andrin, Hessam Bagherinezhad, Hongyu Ren, Hunter Lightman, Hyung~Won Chung, Ian Kivlichan, Ian O'Connell, Ian Osband, Ignasi~Clavera Gilaberte, Ilge Akkaya, Ilya Kostrikov, Ilya Sutskever, Irina Kofman, Jakub Pachocki, James Lennon, Jason Wei, Jean Harb, Jerry Twore, Jiacheng Feng, Jiahui Yu, Jiayi Weng, Jie Tang, Jieqi Yu, Joaquin~Quiñonero Candela, Joe Palermo, Joel Parish, Johannes Heidecke, John Hallman, John Rizzo, Jonathan Gordon, Jonathan Uesato, Jonathan Ward, Joost Huizinga, Julie Wang, Kai Chen, Kai Xiao, Karan Singhal, Karina Nguyen, Karl Cobbe, Katy Shi, Kayla Wood, Kendra Rimbach, Keren Gu-Lemberg, Kevin Liu, Kevin Lu, Kevin Stone, Kevin Yu, Lama Ahmad, Lauren Yang, Leo Liu,
  Leon Maksin, Leyton Ho, Liam Fedus, Lilian Weng, Linden Li, Lindsay McCallum, Lindsey Held, Lorenz Kuhn, Lukas Kondraciuk, Lukasz Kaiser, Luke Metz, Madelaine Boyd, Maja Trebacz, Manas Joglekar, Mark Chen, Marko Tintor, Mason Meyer, Matt Jones, Matt Kaufer, Max Schwarzer, Meghan Shah, Mehmet Yatbaz, Melody~Y. Guan, Mengyuan Xu, Mengyuan Yan, Mia Glaese, Mianna Chen, Michael Lampe, Michael Malek, Michele Wang, Michelle Fradin, Mike McClay, Mikhail Pavlov, Miles Wang, Mingxuan Wang, Mira Murati, Mo~Bavarian, Mostafa Rohaninejad, Nat McAleese, Neil Chowdhury, Neil Chowdhury, Nick Ryder, Nikolas Tezak, Noam Brown, Ofir Nachum, Oleg Boiko, Oleg Murk, Olivia Watkins, Patrick Chao, Paul Ashbourne, Pavel Izmailov, Peter Zhokhov, Rachel Dias, Rahul Arora, Randall Lin, Rapha~Gontijo Lopes, Raz Gaon, Reah Miyara, Reimar Leike, Renny Hwang, Rhythm Garg, Robin Brown, Roshan James, Rui Shu, Ryan Cheu, Ryan Greene, Saachi Jain, Sam Altman, Sam Toizer, Sam Toyer, Samuel Miserendino, Sandhini Agarwal, Santiago Hernandez,
  Sasha Baker, Scott McKinney, Scottie Yan, Shengjia Zhao, Shengli Hu, Shibani Santurkar, Shraman~Ray Chaudhuri, Shuyuan Zhang, Siyuan Fu, Spencer Papay, Steph Lin, Suchir Balaji, Suvansh Sanjeev, Szymon Sidor, Tal Broda, Aidan Clark, Tao Wang, Taylor Gordon, Ted Sanders, Tejal Patwardhan, Thibault Sottiaux, Thomas Degry, Thomas Dimson, Tianhao Zheng, Timur Garipov, Tom Stasi, Trapit Bansal, Trevor Creech, Troy Peterson, Tyna Eloundou, Valerie Qi, Vineet Kosaraju, Vinnie Monaco, Vitchyr Pong, Vlad Fomenko, Weiyi Zheng, Wenda Zhou, Wes McCabe, Wojciech Zaremba, Yann Dubois, Yinghai Lu, Yining Chen, Young Cha, Yu~Bai, Yuchen He, Yuchen Zhang, Yunyun Wang, Zheng Shao, and Zhuohan Li.
\newblock Openai o1 system card, 2024{\natexlab{b}}.
\newblock URL \url{https://arxiv.org/abs/2412.16720}.

\bibitem[Song et~al.(2025)Song, Su, Qu, Zhou, and Cheng]{song2025prmbench}
Mingyang Song, Zhaochen Su, Xiaoye Qu, Jiawei Zhou, and Yu~Cheng.
\newblock {{PRMBench}}: {{A Fine-grained}} and {{Challenging Benchmark}} for {{ Process-Level Reward Models}}, 2025.

\bibitem[Tan et~al.(2025)Tan, Yao, Qu, Li, Yang, Lu, Wang, Qiu, Chu, Xu, and Qi]{tan2025universalprm}
Xiaoyu Tan, Tianchu Yao, Chao Qu, Bin Li, Minghao Yang, Dakuan Lu, Haozhe Wang, Xihe Qiu, Wei Chu, Yinghui Xu, and Yuan Qi.
\newblock Aurora:automated training framework of universal process reward models via ensemble prompting and reverse verification, 2025.
\newblock URL \url{https://arxiv.org/abs/2502.11520}.

\bibitem[Valdenegro-Toro \& Saromo(2022)Valdenegro-Toro and Saromo]{valdenegrotoro2022deeperlookaleatoricepistemic}
Matias Valdenegro-Toro and Daniel Saromo.
\newblock A deeper look into aleatoric and epistemic uncertainty disentanglement, 2022.
\newblock URL \url{https://arxiv.org/abs/2204.09308}.

\bibitem[Wang et~al.(2024)Wang, Li, Shao, Xu, Dai, Li, Chen, Wu, and Sui]{wang2024mathshepherd}
Peiyi Wang, Lei Li, Zhihong Shao, R.~X. Xu, Damai Dai, Yifei Li, Deli Chen, Y.~Wu, and Zhifang Sui.
\newblock Math-{{Shepherd}}: {{Verify}} and {{Reinforce LLMs Step-by-step}} without {{Human Annotations}}, 2024.

\bibitem[Wei et~al.(2024)Wei, Hanning, Nan, and Tong]{xiong2024rlhflowprm}
Xiong Wei, Zhang Hanning, Jiang Nan, and Zhang Tong.
\newblock An implementation of generative prm., 2024.
\newblock URL \url{https://github.com/RLHFlow/RLHF-Reward-Modeling}.

\bibitem[Yang et~al.(2024)Yang, Yang, Zhang, Hui, Zheng, Yu, Li, Liu, Huang, Wei, Lin, Yang, Tu, Zhang, Yang, Yang, Zhou, Lin, Dang, Lu, Bao, Yang, Yu, Li, Xue, Zhang, Zhu, Men, Lin, Li, Xia, Ren, Ren, Fan, Su, Zhang, Wan, Liu, Cui, Zhang, and Qiu]{qwen2.5}
An~Yang, Baosong Yang, Beichen Zhang, Binyuan Hui, Bo~Zheng, Bowen Yu, Chengyuan Li, Dayiheng Liu, Fei Huang, Haoran Wei, Huan Lin, Jian Yang, Jianhong Tu, Jianwei Zhang, Jianxin Yang, Jiaxi Yang, Jingren Zhou, Junyang Lin, Kai Dang, Keming Lu, Keqin Bao, Kexin Yang, Le~Yu, Mei Li, Mingfeng Xue, Pei Zhang, Qin Zhu, Rui Men, Runji Lin, Tianhao Li, Tingyu Xia, Xingzhang Ren, Xuancheng Ren, Yang Fan, Yang Su, Yichang Zhang, Yu~Wan, Yuqiong Liu, Zeyu Cui, Zhenru Zhang, and Zihan Qiu.
\newblock Qwen2.5 technical report.
\newblock \emph{arXiv preprint arXiv:2412.15115}, 2024.

\bibitem[Yuan et~al.(2024)Yuan, Li, Chen, Cui, Ding, Zhang, Zhou, Liu, and Peng]{yuan2024implicitprm}
Lifan Yuan, Wendi Li, Huayu Chen, Ganqu Cui, Ning Ding, Kaiyan Zhang, Bowen Zhou, Zhiyuan Liu, and Hao Peng.
\newblock Free process rewards without process labels.
\newblock \emph{arXiv preprint arXiv:2412.01981}, 2024.

\bibitem[Zhang et~al.(2025)Zhang, Zheng, Wu, Zhang, Lin, Yu, Liu, Zhou, and Lin]{zhang2025lessons}
Zhenru Zhang, Chujie Zheng, Yangzhen Wu, Beichen Zhang, Runji Lin, Bowen Yu, Dayiheng Liu, Jingren Zhou, and Junyang Lin.
\newblock The {{Lessons}} of {{Developing Process Reward Models}} in {{ Mathematical Reasoning}}, 2025.

\bibitem[Zheng et~al.(2024)Zheng, Zhang, Zhang, Lin, Lu, Yu, Liu, Zhou, and Lin]{zheng2024processbench}
Chujie Zheng, Zhenru Zhang, Beichen Zhang, Runji Lin, Keming Lu, Bowen Yu, Dayiheng Liu, Jingren Zhou, and Junyang Lin.
\newblock {{ProcessBench}}: {{Identifying Process Errors}} in {{Mathematical Reasoning}}, 2024.

\end{thebibliography}
\bibliographystyle{colm2025_conference}

\appendix
\newpage
\section{LLM-as-Judger Prompt Template}~\label{sec:judge_prompt}
For LLM-as-Judger, we follow the prompt in \citet{zhang2025lessons}.
\lstset{
	basicstyle=\ttfamily\scriptsize, 
	backgroundcolor=\color{gray!10}, 
	frame=tb, 
	rulecolor=\color{black}, 
	breaklines=true, 
	postbreak=\mbox{\textcolor{red}{$\hookrightarrow$}\space}, 
}
\begin{lstlisting}
I will provide a math problem along with a solution. They will be formatted as follows:
[Math Problem]
<math_problem>
...(math problem)...
</math_problem>
[Solution]
<paragraph_1>
...(paragraph 1 of solution)...
</paragraph_1>
...
<paragraph_n>
...(paragraph n of solution)...
</paragraph_n>

Your task is to review each paragraph of the solution in sequence, analyzing,
verifying, and critiquing the reasoning in detail. You need to provide the
analyses and the conclusion in the following format:
<analysis_1>
...(analysis of paragraph 1)...
</analysis_1>
...
<analysis_n>
...(analysis of paragraph n)...
</analysis_n>
<conclusion>
Correct/Incorrect
</conclusion>

* When you analyze each paragraph, you should use proper verification, recalculation, or reflection to indicate whether it is logically and mathematically valid. Please elaborate on the analysis process carefully.
* If an error is detected in any paragraph, you should describe the nature and cause of the error in detail, and suggest how to correct the error or the correct approach. Once a paragraph is found to contain any error, stop further analysis of subsequent paragraphs (as they may depend on the identified error) and directly provide the conclusion of "Incorrect." For instance, given a solution of five paragraphs, if an error is found in the third paragraph, you should reply in the following format:
<analysis_1>
...(analysis of paragraph 1)...
</analysis_1>
<analysis_2>
...(analysis of paragraph 2)...
</analysis_3>
<analysis_3>
...(analysis of paragraph 3; since an error is found here, also provide detailed critique and correction guideline)...
</analysis_3>
<conclusion>
Incorrect
</conclusion>
Note that the analyses of paragraphs 4 and 5 should be skipped as the paragraph 3 has been found to contain an error.
* Respond with your analyses and conclusion directly.
--------------------------------------------------
The following is the math problem and the solution for your task:
[Math Problem]
{tagged_problem}
[Solution]
{tagged_response}
\end{lstlisting}

\section{More Experiment Results}

\subsection{Problem diversity is important for Training PRMs}~\label{sec:why_not_use_prm800k}

PRM800K~\citep{lightman2023lets} is a widely used and human-annotated dataset for PRM training, which contains 800K step-level labels across 75K tree-of-thoughts solutions to 12K MATH~\citep{hendrycks2021math}. Our empirical results show that models trained on our dataset (100K samples from 100K diverse questions) consistently and significantly outperform those trained on PRM800K\footnote{\url{https://huggingface.co/datasets/HuggingFaceH4/prm800k-trl-dedup}} (369K samples from only 12K questions) on ProcessBench. These findings suggest that problem diversity plays a more crucial role in PRM training than the number of step-level annotations.

\begin{table}[t]
    \centering
\resizebox{\textwidth}{!}{
    \begin{tabular}{cccc}
    \toprule
         & \textbf{\# Problem set} & \# \textbf{CoT Trajectories} & \textbf{ProcessBench F1 score} \\
    \midrule
    \textbf{PRM800K}   & 7,500 & 460,000 & 0.575 \\
    \textbf{NuminaMath (Random Selected)} & 100,000 & 100,000 & 0.673 \\
    \bottomrule
    \end{tabular}}
    \caption{Comparison between PRM800K and 100K data collected from NuminaMath labeled by Qwen-QwQ.}
    \label{tab:why_not_use_prm800k}
\end{table}
\section{Annotation Cost Estimation}~\label{sec:labeling_cost_estimation}
We estimate the labeling cost based on the statistics of our 1M data collected from NuminaMath~\cite{numinamath} using Qwen2.5-Math-7B-Instruct and Qwen2.5-Math-72B-Instruct. We introduce the statistics in Table~\ref{tab:data_statistics}.

\begin{table}[h]
    \centering
    \resizebox{0.9\textwidth}{!}{
    \begin{tabular}{llc}
    \toprule
    & \textbf{Value} & \textbf{Source} \\
    \midrule
    \textbf{\# Reasoning Steps (\(S\))} & 8.845 & Qwen Models' rollouts\\
    \textbf{\# Tokens per Rollout (\(R\))} & 625.098 & Qwen Models' rollouts \\
    \textbf{\# Tokens per Critic Response from Judge (\(C\))} & 1,919.860 & Qwen-QwQ's responses as LLM-as-Judge\\
    \bottomrule
    \end{tabular}}
    \caption{Statistics of 1M NuminaMath CoT Trajectories collected by Qwen2.5-Math Models.}
    \label{tab:data_statistics}
\end{table} 

In addition to the statistics, we also use \(N\) to denote the data number of the dataset and show this statistic of each model's training dataset in Table~\ref{tab:data_number}

\begin{table}[h]
    \centering
    \resizebox{0.7\textwidth}{!}{
    \begin{tabular}{ll}
    \toprule
    \textbf{Dataset} & \textbf{\# Labeled Data} \\
    \midrule
    \textbf{\ourmethod} & 624,000 (labeled in two stages) \\
    \textbf{Qwen2.5-Math-PRM-Math-shepherd} & 860,000 \\
    \textbf{Qwen2.5-Math-PRM} & 860,000 \\
    \textbf{UniversalPRM} & 690,000 \\
    \bottomrule
    \end{tabular}}
    \caption{Data number of datasets.}
    \label{tab:data_number}
\end{table} 

Using the statistics, we compute the estimated labeling cost for \ourmethod, Qwen2.5-Math-PRM-Math-shepherd~\citep{zhang2025lessons}, Qwen2.5-Math-PRM\citep{zhang2025lessons}, UniversalPRM~\citet{tan2025universalprm} as follows:

\begin{itemize}
    \item Qwen2.5-Math-PRM-Math-shepherd: \(N \times S \times 8 \times R / 2\), where \(8\) is the number of rollouts per step set in \citet{zhang2025lessons}. We divided by two since the number of tokens for rollouts varies based on the position of reasoning step. For latter reasoning step, it requires less reasoning tokens. As a result, the expectation of tokens per rollout should be half of the number of tokens of the complete rollout.
\item Qwen2.5-Math-PRM: \(N \times S \times 8 \times R / 2 + N * C\). It used consensus filtering for each data, the cost is both from MathShepherd (\(S \times 8 \times R / 2\)) and LLM-as-Judge (C).
\item UniversalPRM: \(N \times C \times 4 + N \times S\), where 4 is the number of ensemble prompts from the original paper~\citep{tan2025universalprm} and another \(N \times S\) is for its semantic-based step seperation.
\item \ourmethod: \(N \times C\). We solely use Qwen-QwQ as Judge and do not include any other operations.
\end{itemize}

\end{document}